\documentclass[letterpaper]{article} % DO NOT CHANGE THIS
\usepackage{aaai23}  % DO NOT CHANGE THIS
\usepackage{times}  % DO NOT CHANGE THIS
\usepackage{helvet}  % DO NOT CHANGE THIS
\usepackage{courier}  % DO NOT CHANGE THIS
\usepackage[hyphens]{url}  % DO NOT CHANGE THIS
\usepackage{graphicx} % DO NOT CHANGE THIS
\usepackage{enumitem}
\usepackage{amsmath}
\usepackage{subfig}
\usepackage{bm}
\usepackage{amsfonts}
\usepackage{wrapfig}
\usepackage{natbib}  % DO NOT CHANGE THIS AND DO NOT ADD ANY OPTIONS TO IT
\usepackage{caption} % DO NOT CHANGE THIS AND DO NOT ADD ANY OPTIONS TO IT
\usepackage{algorithm}
\usepackage{algorithmic}
\newcommand{\tabincell}[2]{\begin{tabular}{@{}#1@{}}#2\end{tabular}}
\usepackage{mathtools}
\usepackage{amsthm}
\theoremstyle{plain}

\theoremstyle{definition}

\theoremstyle{remark}

\newcommand\Mark[1]{\textsuperscript#1}
\usepackage{newfloat}
\usepackage{listings}
\DeclareCaptionStyle{ruled}{labelfont=normalfont,labelsep=colon,strut=off} % DO NOT CHANGE THIS
\floatstyle{ruled}
\newfloat{listing}{tb}{lst}{}
\floatname{listing}{Listing}
\title{Bellman Meets Hawkes:  Model-Based Reinforcement Learning via Temporal Point Processes}
\author{
    Chao Qu\equalcontrib\Mark{1},
    Xiaoyu Tan\equalcontrib\thanks{Corresponding author. txywilliam1993@outlook.com}\Mark{1},
    Siqiao Xue\Mark{1}, 
    Xiaoming Shi\Mark{1},
    James Zhang\Mark{1},
    Hongyuan Mei\Mark{2}
}
\affiliations{
    \Mark{1} Ant Group, Hangzhou, China. \\
    \Mark{2} Toyota Technological Institute at Chicago, Chicago, IL, United States.\\
    chaoqu.technion@gmail.com,txywilliam1993@outlook.com, \{siqiao.xsq,peter.sxm,james.z\}@antgroup.com\\
    hongyuan@ttic.edu
}

\usepackage{bibentry}
\begin{document}

\maketitle

\begin{abstract}
We consider a sequential decision making problem where the agent faces the environment characterized by the stochastic discrete events and seeks an optimal intervention policy such that its long-term reward is maximized. This problem exists ubiquitously in social media, finance and health informatics but is rarely investigated by the conventional research in reinforcement learning. To this end, we present a novel framework of the model-based reinforcement learning where the agent's actions and observations  are  asynchronous stochastic discrete events occurring in continuous-time.  We model the dynamics of the environment by Hawkes process with external intervention control term and develop an algorithm to embed such process in the Bellman equation which guides the direction of the value gradient. We demonstrate the superiority of our method  in both synthetic simulator and real-data experiments. 
\end{abstract}

\section{Introduction}

The last several years have witnessed the great success of reinforcement learning (RL) including the  video game playing \citep{mnih2015human}, robot manipulation \citep{gu2017deep}, autonomous driving \citep{shalev2016safe} and many others \citep{lazic2018data,dalal2016hierarchical}. Most of them focus on the problem where the system of interest evolves continuously with time, e.g., a trajectory of a robot arm.  

%For instance, the dynamic of joints in the robot follows the physical law of motion.

\begin{figure*}
\centering
\includegraphics[width= 0.9\textwidth]{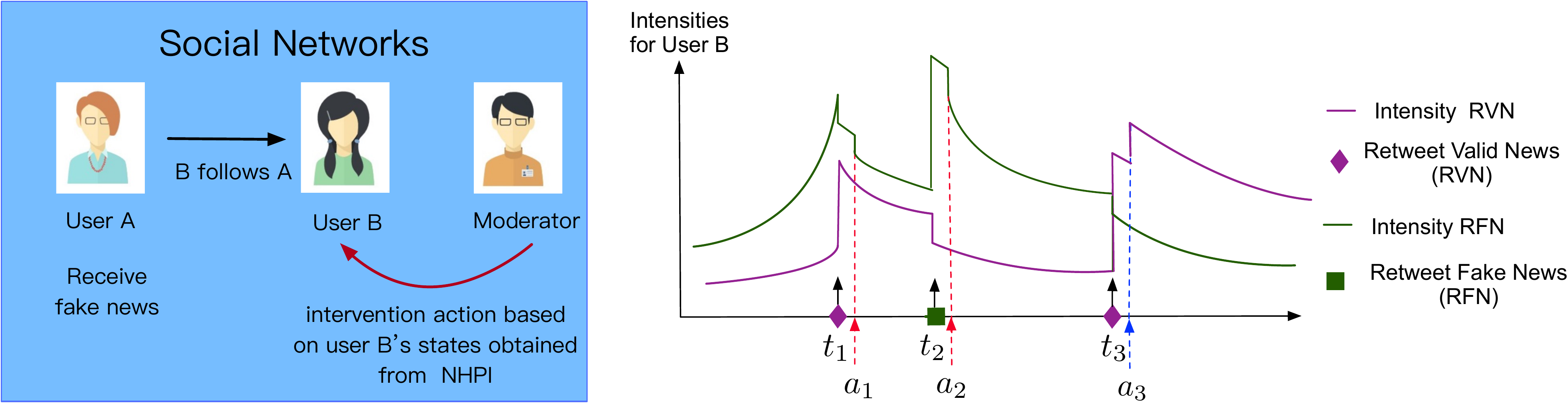}

\caption{The figure in the left panel illustrates the fake new mitigation problem. The user B may follow the fake news from user A and spread it to others. The moderator may expose valid news to user B to incentivize more spontaneous mitigation events (e.g., valid news at $t_3$). The right panel is the corresponding intensity function of valid news  and the fake news  modeled by the Neural Hawkes Process with intervention. 
The probability that an event occurs in the infinitesimal interval $[t, t+dt]$ is determined 
by intensity times $dt$ (i.e., $\lambda_i(t)dt$) as that in TPP. 
Events of valid news and fake news inhibit each other. When events occur 
at time $t_1,t_2,t_3$, there are immediate effects (the sudden jumps)
and long-term effects (decay with time) in intensity. In addition, the agent has two types of 
actions to intervene in the event stream: red action to inhibit fake news event and blue one to excite valid news event, which cause instant effects on the intensity of events. In our setting, the agent takes 
actions at timestamp $t_i$ or with a small delay (see the section of problem setup ).}\label{fig:problem_setting}
\vspace{-6mm}
\end{figure*}

However, the conventional research in RL may omit a category of system that evolves continuously and may be interrupted by  stochastic events \emph{abruptly} ( See the jumps at $t_1,t_2,t_3$ in Figure \ref{fig:problem_setting}). Such system exists ubiquitously  in the social and information science and therefore necessitates the research of reinforcement learning in these  domains to extend its applicability in the real-world problems \citep{farajtabar2017fake,wang2018stochastic}, in which the agent seeks an optimal intervention policy so as to improve the future course of events. Concrete examples may include:

\begin{itemize}[leftmargin=*,nolistsep,nosep]
    \item Social media. Social media website allows users to create and share content. Retweet can form as users  resharing and broadcasting others’ tweet to their friends and followers. Such stochastic events  would steer the behaviors of other tweet users \citep{rizoiu2017tutorial}; the platform (agent) may want to seek a policy to effectively mitigate the fake news by optimizing the performance of real news propagation over the network \cite{farajtabar2017fake}.
    
    \item Medical events. The stream of medical events would include the blood test, diagnosis, treatment and so on. To cure the patients, a doctor (the agent) often seeks the optimal treatments  based on the clinical conditions. 
\end{itemize}

 In the above problems, given the observed trajectories of the event sequences in continuous-time, the agent can impose the intervention  on the environment  to maximize the long term reward. For instance, the platform can post the valid news in the social media to steer the spontaneous user mitigation activities of fake news. In the RL context, the action is to post valid news and the observation is the users' following behaviors such as retweet while the reward can be quantified by event exposure counts. However, the following characteristics of event sequences bring new challenges: 
 \begin{enumerate}[leftmargin=*,nolistsep,nosep]
     \item The inter-event time intervals are \emph{not} constant while conventional RL considers the synchronized action and observations. For instance, the healthy patient with less severe symptoms  would visit doctors occasionally, while the unhealthy patients receive the test and treatment more frequently. Thus it is crucial  to predict both which events are likely to happen next and when they will happen. Directly adapting conventional RL algorithm to its Semi-MDP (SMDP) version can not address this issue well (see our \emph{ablation study} in Appendix. )
     \item How to describe the dynamics of an abrupt event if the agent imposes intervention? Note that these events may help  cause or prevent future events in a complex mechanism, e.g., in Figure \ref{fig:problem_setting}, the platform  posts the valid news to steer the spontaneous user activities.   In general, the valid news and fake news can inhibit each other. Specifically, the dynamic system characterized by continuous trajectories will be abruptly changed by discrete events. However, the dynamic models of most model-based RL methods are continuous \emph{without} endogenous jumps (e.g., the jump corresponding to the spontaneous user activities), and therefore are not suitable for learning the continuous and discrete dynamics of such hybrid system.  
 \end{enumerate}

 To this end, we propose a novel model-based reinforcement learning, where the agent seeks the long term reward by imposing the interventions (event) on this hybrid system.  In our setting,  both the actions taken by the agent and the observations from the environment are asynchronous stochastic event in continuous time.   To describe the dynamics of the environment, we adapt \underline{N}eural \underline{H}awkes \underline{P}rocess \citep{zhang2020self,zuo2020transformer,mei2017neural} to the case with exogenous \underline{I}nterventions (i.e., the actions in RL) and name it NHPI for short.  Instead of discretizing time into fixed-width intervals in conventional RL, we treat timestamps as random variables. The latent representations of NHPI implicitly describe when and which type of event will happen in the future by learning the intensity function $\lambda$. We embed  the latent space of the learned dynamic model into the Bellman equation and derive a  stochastic value gradient algorithm in the semi-Markov decision process to cope with the challenge of irregular timestamps.

It is known that Hawkes process can be reformulated as stochastic differential equations (SDEs) \emph{with endogenous jumps} caused by stochastic events inside the environment itself. The dynamic model of most traditional model-based RL methods can be described by SDEs or ODEs  \emph{without} endogenous jumps (if we adapt them in the continuous-time setting). Thus, it reveals the intrinsic difference between our approach and the previous model-based RL methods, which is evidenced by the experimental results showing that our algorithm outperforms model-free and model-based RL baselines  with a notable margin by considering the  discontinuity nature of the jumps. We discuss the connections to ODE approach  and world model \cite{hafner2019dream} thoroughly in Appendix.

\textbf{Contributions:}  We propose a model-based RL algorithm to solve the problem that can evolve both continuously and can be interrupted by the stochastic event abruptly.  Our method captures the jump and the irregular time-interval property in the systems and  extends the applicability of RL in the area of social media, medical events and many others. Empirically, we demonstrate that our method surpasses the performances of other state-of-the-art model-free and model-based RL algorithms. We release our code at \url{https://github.com/WilliamBUG/Event_driven_rl/tree/main}

\section{Related work}
%  We  discuss related literature on the temporal point processes and reinforcement learning. 

Temporal point processes are prevalent in modeling the sequences of events happening at irregular intervals and have wide applications in e-commerce systems, social media, and traffic prediction \citep{hawkes1974cluster,daley2007introduction}. Recent works combine temporal point process with deep learning  and design algorithms to learn complex behavior from real-world data \citep{mei2017neural,du2016recurrent,zuo2020transformer,zhang2020self,xue2021graphpp,xue2022hypro}. However, these works just focus on the \emph{prediction} problem rather than the \emph{control} problem, i.e., the agent does not make any decision to impose the intervention on the environment. The only exceptions are \cite{wang2018stochastic,zarezade2017steering,upadhyay2018deep,farajtabar2017fake}.   The problem is formulated into  SDEs in \cite{wang2018stochastic,zarezade2017steering} and solved with the optimal control theory.  However they assume that the dynamic are known, which is not applicable in most real-world problems. TPPRL  proposes a modified version of REINFORCE algorithm \cite{upadhyay2018deep}. In general, such simple model-free algorithm suffers from high sample-complexity. The authors in \cite{farajtabar2017fake} use linear least square temporal difference (LSTD) learning to learn a policy to mitigate the fake news over the social media. However, it can only apply to simple problem due to the limitation of LSTD.  Our work builds a flexible model to describe the underlying dynamics of the event stream intervened with the action, and then we propose a sample-efficient model-based RL algorithm where the agent seeks the optimal policy  to achieve some desired states and to maximize the long-term reward. 

There are a few works in literature on the Semi-MDPs and the continuous-time reinforcement learning
\citep{munos2006policy,bradtke1995reinforcement}, which are natural formalism  for the problem with irregular time intervals. However, these works are generally model-free algorithm and have low sample efficiency. The problem in social science is  partially observable in the real-world, which made RNN and its variants popular to summarize the long-term dependencies and describe the belief states in RL \citep{oh2015action}.  In model-based approaches \cite{ha2018world,hafner2019dream,hafner2019learning}, a world model is often learned from the past experience and generates new imaginary data to feed the agent.  These algorithms also have downsides, i.e., they generally discretize the time evenly. The authors in \cite{du2020model,yildiz2021continuous} use neural ODE to create the world model and hence can cope with the continuous-time setting. However their work can not handle the hybrid dynamic with jumps and therefore their empirical results focus on the Openai gym (e.g., swimmer) \cite{brockman2016openai}. 
As shown in Appendix, our model can be thought of as SDEs with jumps. To the best of our knowledge, none work mentioned above (except ours) considers the dynamics with jumps sparked by the event, which is essential in modeling real-world problems, since such jump term related to the counting measure reflects the nature in the social science, epidemic process and many others \citep{cox1980point,rasmussen2018lecture}. In our experimental result, we also demonstrate our algorithm outperforms the state of the arts.

\section{Preliminary}\label{section:preliminary}
As we discussed in the introduction, we need to properly handle the problem with irregular time interval which is caused by the abrupt event. To this end, we introduce temporal point process and the Semi-MDPs, which are building blocks of our work.

\textbf{Temporal Point Process:} A temporal point process (TPP) is a random process whose realization consists of a list of discrete events localized in time $\mathcal{T} = \{t_1,...,t_N\}$, where $t_i< t_{i+1}$ \citep{daley2007introduction}. 
  The marked temporal point process is a stochastic process whose realization consists not only the event time but also event type   $k\in \mathcal{K}$, where $ \mathcal{K}:= \{1,...,K\}$.   Let $Es= \{(t_i,k_i)\}_{i=1}^L $ be an event sequence with length $L$, where $k_i$ is the $i$-th event of the sequence $Es$ and $t_i$ is the timestamp of the $i$-th event. Then the history is defined  as  $\mathcal{H}_t:= \{ (t_i,k_i)| t_i< t, k_i\in \mathcal{K}\}$.
  In practice, $k_i$ can represent the messages posted in the social media, or the treatment given by the doctor while $t_i$ is the  corresponding occurring time of such event. We use $\lambda_k(t)$ to represent the rate of an event with type $k$ and denote $\lambda(t) = \sum_{k=1}^{K}\lambda_k(t)$ as the total intensity of all events. Hawkes process  captures the \emph{interactions} between different types of event, in which past events conspire to raise the intensity of each type of event \citep{hawkes1974cluster}. In particular, the intensity with type $k$ is modeled as:$\lambda_k(t) = \mu_k +\sum_{h: t_h<t}\beta_{k_h,k} \exp (-\zeta(t-t_h))$ where $\mu_k$ is the base intensity of event type $k$, $\beta_{j,k}\geq 0 $ is the degree to which an event of type $j$ excites type $k$, and $\zeta>0 $ is the decay rate of that excitation.  Random sequence can be generated from the model by the thinning algorithm \citep{lewis1979simulation}. Recent works \citep{mei2017neural,zuo2020transformer,zhang2020self} combine the progress in  deep learning to enhance the capacity of this process, e.g., using RNN or transformer to approximate the intensity function  and can predict online and off-line human actions accurately in social media.

\textbf{Semi-MDPs:} A semi-Markov decision process (SMDP) is a tuple $(\mathcal{S},\mathcal{A},P,\mathcal{R},\mathcal{T},F,\rho)$, where $\mathcal{S}$ is the state space, $\mathcal{A}$ is the action space, $\mathcal{T}$ is a transition time space, $\mathcal{P}$ is the transition probability, and $e^{-\rho}$ is the discount factor with $\rho>0$~\citep{bradtke1995reinforcement}. A Semi-Markov process evolves as follows. The next state $s'$ is chosen according to the transition probabilities $P(s'|s,a)$. Conditioned on the event that the next state is $s'$, the time interval $\tau \in \mathcal{T}$ between the $s$ to $s'$  has the probability distribution of $F(\cdot|a,s,s')$. The reward function $r(t):= R(s(t),a(t))$. The objective of the agent is to find a policy $\pi(a|s)$ to maximize the expected infinite horizon discounted reward $\mathbb{E} [\int_{0}^{\infty} e^{-\rho t}r(t)dt  ]$. The Bellman equation for the SMDPs is  defined as follows \citep{bradtke1995reinforcement}.
\begin{equation}\label{equ:bellman_equation}
\begin{aligned}
     V(s)= & \sum_{a,s'}\pi(a|s)P(s'|a,s) [\int_{0}^{\infty} \int_{0}^{t}e^{-\rho\tau}r(\tau)d\tau dF(t|s,a,s')\\
     &+   \int_{0}^{\infty} e^{-\rho t} V(s')dF(t|s,a,s')].
\end{aligned}
\end{equation}

In contrast to the constant time interval in MDP,  $\tau$ in SMDP is stochastic \cite{sutton1999between}.
 Our high level idea is to model such interval with the temporal point processes so as to describe the dynamics of the control problem in SMDPs.
We  provide some concrete examples  in the social media problem to relate these definitions to the real world problems in the following. The state $s$ is the count of the events with different types, e.g., the numbers of fakes and real news received by the users, or the embedding of these events sequence in the \emph{hidden space} of RNN.  The action $a$ is the intervention, e.g., the network moderator  exposes different types of the real news (actions) to users to match their exposure to fake news.  The reward can be the distance between the target frequency of the event and the current frequency, and the cost of different actions can be different. 
\section{Our method}

 This section is organized as follows: we first describes the setting of the problem. Then we modify the Hawkes process to incorporate the effect of actions.  Using NHPI, we derive a model-based stochastic value gradient algorithm for RL  in the SMDPs setting. We relate this  model to the SDEs with jumps and discuss the key difference between our reinforcement learning algorithm and traditional RL through our SDE reformulation in Appendix.

\subsection{Problem setup} \label{section:Problem_setup} 

In many problems, including social media,  medical events and many others, the agent observes a steam of the asynchronous stochastic discrete events occurring in continuous-time, such as the user's posts, clicks or likes.  We denote the observations from the environment  $\mathcal{H}_{\leq  t_i}$ up to $t_i$ (including $t_i$) as $\{(t_1,k_1),...,(t_i,k_i)\}$. In our setting, the agent can just intervene in the environment or change its actions $a_i$ at  timestamp $t_i$. We denote its action sequence before time $t_i$ (not including $t_i$) as $a_{<t_i}$. At timestamp $ t_i$, the agent samples its action from a policy, i.e.,   $a_i\sim \pi(\cdot| \mathcal{H}_{\leq t_i},a_{<t_i} )$. Unless otherwise specified, we assume $\mathcal{A} \subseteq \mathcal{K}$, i.e., the action set is a subset of the total event set. After imposing the intervention on the environment, the agent obtains a reward $r(t)$ and aims to maximize the discounted long-term reward,
$\mathbb{E} [\int_{0}^{\infty} e^{-\rho t}r(t)dt  ],$
where the expectation is over the randomness of the transition, time interval and reward in the SMDPs. In practice, the agent may apply the action $a_i$ after  $t_i$ with a small delay $\delta t$ (see the illustration in Figure \ref{fig:problem_setting}). For instance, after realizing that rumors spreads over the social media, the network moderator may need some response time $\delta t$ to apply the strategy to block the diffusion. The doctor provide the treatment for the patient after the medical test. This $\delta t$ can be arbitrary predefined positive value in accordance with tasks. In our experiment, we impose the intervention at the next coming clock tick for simplicity.  To ease our notion here, we simply assume $a_i$ occurs at time $t_i$. 

%\xue Also remark that our setting is slightly different from \citet{upadhyay2018deep}, in which it assumes that the agent can impose the intervention at any time. 

\subsection{NHPI Model }\label{section:model_learning}
\begin{figure}
	\begin{center}
		\includegraphics[width=0.48\textwidth]{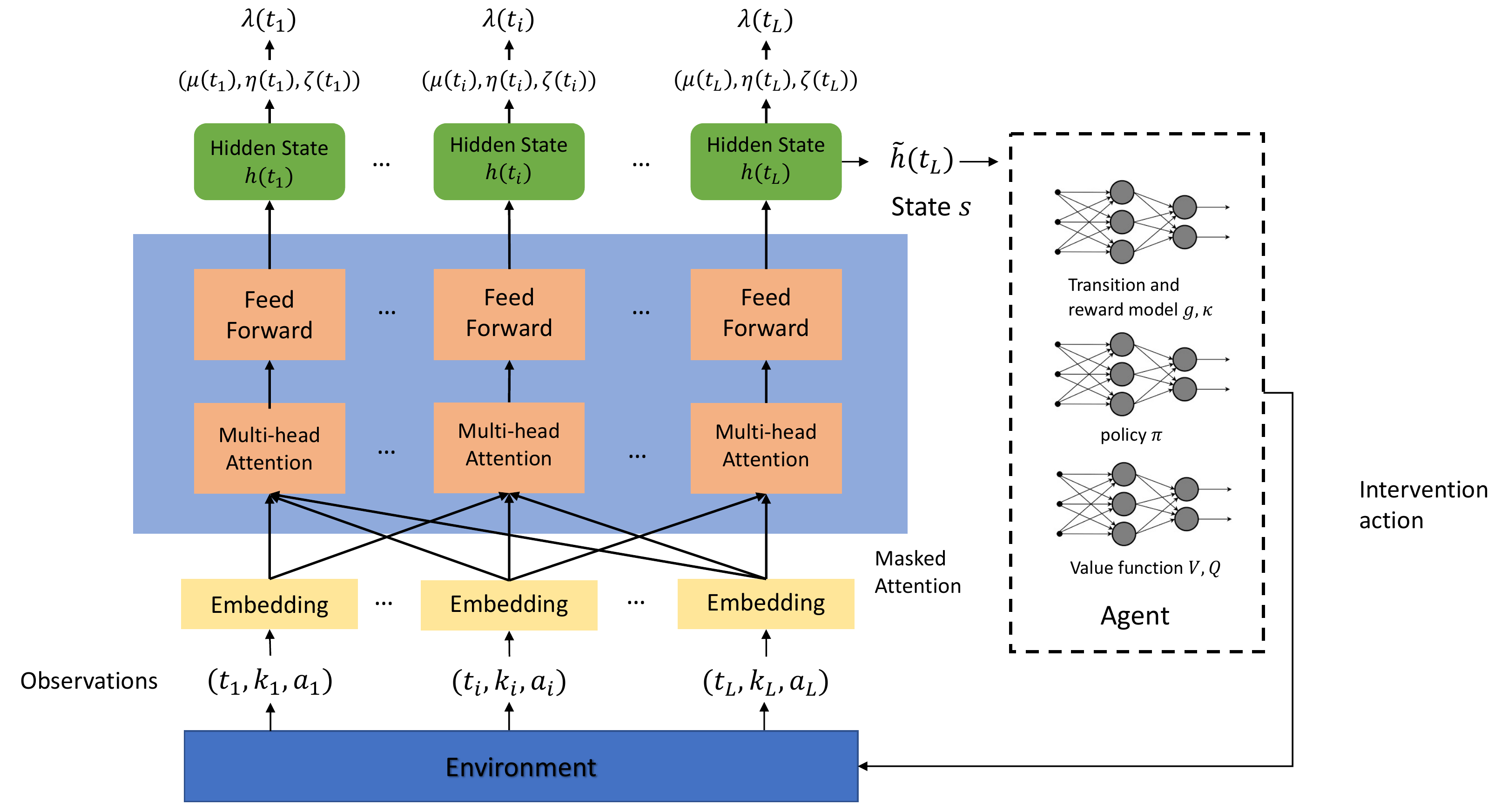}
	\end{center}
\vspace{-3mm}
\caption{Structure of NHPI. We embed the $(t_i,k_i,a_i)$ into a dense vector, and pass two multi-head attention+ feed forward modules with masks. The outputs are $\mathbf{h}(t_i),\bm{\mu}(t_i),\bm{\eta}(t_i),\bm{\zeta}(t_i),\bm{\lambda}(t_i)$. We then feed states, i.e., the latent representation $\bm{\Tilde{h}}$ to our RL module.}
\label{fig:NIHP}
\vspace{-4mm}
\end{figure}

Recall that our high level idea is to leverage the Hawkes process to model  \emph{time intervals} $\tau$ in SMDPs so as to describe the underlying dynamics of stochastic event.
In the following, we demonstrate how to modify the neural Hawkes process to the RL setting. In particular, we build a dynamic model following the recent progress in attention \cite{vaswani2017attention} and neural Hawkes process \cite{mei2017neural,zhang2020self,zuo2020transformer}. It is natural to extend the definition of the event stream in the section of preliminary to the event-action stream $Es=\{(t_i,k_i,a_i)\}$ to incorporate exogenous intervention $a$. The whole structure is summarized in Figure \ref{fig:NIHP}, where the latent variable $\bm{h}(t_i)$ is a key vector which summarizes the history and decides the intensity $\lambda(t_i)$ of the Hawkes process. Then the latent state feeds into the agent as the state information. Besides the policy function and value function, the agent builds the transition and reward model over the latent state .

% \begin{figure}
% 	\begin{center}
% 		\includegraphics[width=0.48\textwidth]{Figure/NHPI.pdf}
% 	\end{center}
% \vspace{-3mm}
% \caption{Structure of NHPI. We embed the $(t_i,k_i,a_i)$ into a dense vector, and pass two multi-head attention+ feed forward modules with masks. The outputs are $\mathbf{h}(t_i),\bm{\mu}(t_i),\bm{\eta}(t_i),\bm{\zeta}(t_i),\bm{\lambda}(t_i)$. We then feed states, i.e., the latent representation $\bm{\Tilde{h}}$( See \ref{section:planning_latent} ) to our reinforcement learning module.}
% \label{fig:NIHP}

% \end{figure}

\emph{Encode observations to hidden state $\mathbf{h}$}: As the common practice in RL \cite{ha2018world} and TPP \cite{mei2017neural}, we encode the observations, i.e. the sequence of event types, timestamps and actions $\{(t_i,k_i,a_i)\}_{i=1}^{L}$  into dense embedding $\mathbf{X}\in \mathbb{R}^{L\times N}$ in figure \ref{fig:NIHP}. Then such embedding passes through Multi-head attention, $\mathbf{\Lambda} = softmax(\frac{\mathbf{Q} \mathbf{K}^T}{\sqrt{M_K}})\mathbf{V}$
where $\mathbf{Q}=\mathbf{XW}^Q$, $\mathbf{K}=\mathbf{XW}^K, \mathbf{V}=\mathbf{XW}^V$ are the query, key and value matrix respectively as that in \citep{vaswani2017attention,zuo2020transformer}. $\mathbf{W}^Q,\mathbf{W}^{K}\in \mathbb{R}^{N\times M_{K}}$, $\mathbf{W}^V\in \mathbb{R}^{N\times M_V}$ are learned weight matrix. To avoid observing the future, the attention is equipped with masks. When computing the attention $\mathbf{S}(j,:)$, we mask all the future position.  Then we feed the self-attention $\mathbf{\Lambda}$ into a feed-forward neural network to generate the hidden representation $\mathbf{H}$. The hidden representation at time step $t_i$ is the $i$-th column of $\mathbf{H}$, i.e.,  $\mathbf{h}(t_i) = \mathbf{H}(i,:)$. We defer the description of how to construct embedding $\mathbf{X}$ and parameterizations of neural networks to Appendix. 

\emph{Construct intensity function $\lambda_k(t)$:} We follow  the work in \citep{zhang2020self} to define the intensity function. $  \lambda_k(t) = softplus(\mu_{k}(t_i)+ \big(\eta_{k}(t_i)-\mu_{k}(t_i)\big)\times\exp\big(-\zeta_{k}(t_i)\times(t-t_i)\big)$, where 
$\mu_{k}(t_i) 
 = Relu(\mathbf{w}^T_{\mu,k}\mathbf{h}(t_i)),\eta_{k}(t_i)= Relu (\mathbf{w}^T_{\eta,k} \mathbf{h}(t_i)) $, $ \zeta_{k}(t_i) = softplus(\mathbf{w}^T_{\zeta,k} \mathbf{h}(t_i) )$ and  $\mathbf{w}^T_{\mu,k},\mathbf{w}^T_{\eta,k}, \mathbf{w}^T_{\zeta,k} $ are learned vectors.  Notice that now the intensity also depends on the historical action sequence.

\subsection{ Training NHPI Model}

Given NHPI, the next step is to define the training objective. In the following, we provide the likelihood of the model defined as  the  density function of $(t_i, k_i)$ given the intervention $a_i$ in the event stream $ \{(t_i,k_i,a_i )\}_{i=1}^{L} $ with $t_L<T$.  The Hawkes process has a compact form of the log-likelihood function which only depends on $\lambda$. The log-likelihood of Hawkes process given observations in $[0,T)$ is 
$\sum_{i: t_i<T}\log \lambda_{k_i}(t_i)-\int_{t=0}^T \lambda (t)dt,$
where $\lambda(t)$ is the sum intensity of all event types, i.e., $\lambda(t) = \sum_{k=1}^K \lambda_k(t)$ \citep{rasmussen2018lecture}. We show  NHPI has the same form of the log-likelihood function by adapting the proof in  Hawkes process to allow external action  and include the proof  in Appendix for the completeness.  The term $\sum_{i: t_i<T}\log \lambda_{k_i}(t_i)$  is easy to obtain from samples, since $ \lambda_k(t_i), k\in \{1,...,K\}$ are the  outputs from our model. Specifically, we need to conduct a Monte-Carlo sampling to estimate the integral term and we will defer the details to appendix. Thus the whole objective function can be optimized by stochastic gradient descent.

\subsection{Planning over the latent space of NHPI}\label{section:planning_latent}

\textbf{State}: Recall that $\mathbf{h}(t_i)$ encodes the history  of events and actions $\{(t_j, k_j ,a_j)\}_{j=1,..,i},$ e.g., the history of fakes/real news retweeted by the user and mitigation valid news posted by the platform, which is generally be used as belief state in RL \cite{hafner2019dream}. We define another vector $\mathbf{\Tilde{h}}(t_i)$, which encodes the same history but toggle the last action $a_i = \mathbf{0}$. It is an embedding of history before  an action  at timestamp $t_i$  (i.e., $\mathcal{H}_{\leq t_i},a_{<t_i}$). In our current algorithm, we directly treat $\mathbf{\Tilde{h}}(t_i)$ as  our state $s(t_i)$, i.e., $s(t_i): =\mathbf{\Tilde{h}}(t_i)$ but other transformations on  $\mathbf{\Tilde{h}}$ can also be accommodated.
% e.g., $s(t_i) = f(\mathbf{\Tilde{h}}(t_i))$, where $f$ can vary according to applications.

\textbf{Transition learning}: Given NHPI, we can obtain state $s$ (i.e., $\mathbf{\Tilde{h}}$) at every timestamp. Then we use a neural network $g$ to learn the transition
such that $s(t_{i+1}) \approx g( s(t_{i}),a_{i}, \epsilon(s(t_i)),$
 where $\epsilon$ is the noise term to represent the stochasticity in the transition $P(s(t_{i+1})|s(t_{i}),a_{i})$ of SMDP . In particular, we minimize the error
 \begin{equation}\label{equ:transition_learning}
     \|s(t_{i+1}) -g( s(t_i),a_i, \epsilon(s_{t_i}))\|_2^2,
 \end{equation}

  where $g$ is  a  neural network with  reparameterization trick \citep{kingma2013auto} to incorporate the noise term $\epsilon(s(t_i))$. For instance, it generates the mean and standard deviation of a Gaussian distribution. 
  
\textbf{Reward function learning}: We minimize the error 
    \begin{equation}\label{equ:reward_learning}
       \| r(t_i)- \kappa\big( s(t_i),a_i, \xi(s(t_i) \big)\|_2^2,
    \end{equation}
where $r(t_i)$ is the true reward at timestamp $t_i$, e.g., the event exposure counts.  $\xi$ is Gaussian noise and $\kappa$ is a neural network with reparameterization trick.

\textbf{Value function learning}: Let $\tau_i$ be the time interval between $t_{i}$ and $t_{i+1}$, i.e., $\tau_i= t_{i+1}-t_{i}$.  In SMDPs, the target function for $V(s(t_i))$
is 
$\hat{V}\big(s(t_i)\big) :=  \frac{1-e^{-\rho \tau_i}}{\rho}r(t_i)+e^{-\rho \tau_i} V\big(s(t_{i+1})\big).$ Such target function can be derived from \eqref{equ:bellman_equation}, in particular,  a sampled version of Bellman equation in SMDPs. We defer derivations to Appendix \ref{app_section:value_function_learning}.  Comparing with the temporal difference learning in MDPs \cite{sutton1998introduction}, we have an additional stochastic term $\tau$  corresponding to the inter-event time interval. Recall that we have already obtained the NHPI model, the transition and reward model, and therefore we can  generate imaginary state $s(t_{i+1})$ and reward $r(t_i)$ easily. The time interval $\tau_i$ can be simulated by NHPI with Thinning algorithm \citep{ogata1981lewis}, and thus the target function defined above  can be easily obtained in practice.  Remark that explicitly modelling of $\tau$ with NHPI is one \emph{key} advantage of our method over the conventional RL approaches \cite{haarnoja2018soft,heess2015learning}. While conventional approach  can encode the timestamp using RNN naively,  we can capture the stochasticity of  the time interval $\tau$ much better by TPP. It is evidenced by our experiment, where the inputs of all algorithms are the same, i.e., the sequence of the observation $\{(t_i,k_i,a_i)\}_{i=1}^{L}$, while our method outperforms the conventional RL a lot.  The term $ (1-e^{-\rho \tau_i})/\rho$ is the accumulated discount factor of the interval. We approximate the value function $V$ by a neural network $V_\phi$ and minimize the following error term w.r.t $\phi$.
\begin{equation}\label{equ:value_learning}
    \|V_{\phi}(s(t_i)) - \hat{V}_\phi(s(t_i))\|_2^2.
\end{equation} 
Please note that when we optimize above equation, the target value $\hat{V}_\phi$ is fixed, as a common practice in RL community \citep{haarnoja2018soft}.

\textbf{Policy learning:}  We apply the stochastic policy so that the agent can explore the environment thoroughly \cite{haarnoja2018soft}. In particular, we parameterize the policy $\pi(s(t_i))$ by a neural network $\pi_{\theta}(s(t_i))$. If the action space is discrete, we reparameterize the output by the Gumbel-softmax trick \citep{jang2016categorical}; If it is continuous, we reparameterize the output by the Gaussian distribution \citep{kingma2013auto}. Again, we sample the Bellman equation \eqref{equ:bellman_equation} but with the learned dynamics and value function (see derivations in Appendix \ref{app_section:policy_learning}) as that in SVG \citep{heess2015learning} and Dreamer \citep{hafner2019dream}. Specifically, we replace $s(t_{i+1})$ by  $ g(s(t_i), \pi(s(t_i)), \epsilon(s(t_i)))$, replace $r(t_i)$ by learned reward $\kappa(s(t_i), \pi(s_i),\xi(s(t_i)))$, replace action by $\pi_\theta(s(t_i))$ and rollout Bellman equation one step to obtain
 \begin{equation}
     \begin{aligned}
         V_{\phi,\theta}(s(t_i))= (1-e^{-\rho\tau_i})  \kappa\big(s(t_i), \pi_{\theta}(s(t_i)),\xi(s(t_i))\big) /\rho\\
    + e^{-\rho \tau_i}V_\phi\bigg(g\big(s(t_i), \pi_\theta(s(t_i)),\epsilon(s(t_i))  \big) \bigg). 
     \end{aligned}
 \end{equation}

We can see $V_{\phi,\theta}$ is a function of policy $\pi_\theta$. We do one policy improvement step as that in SVG \citep{heess2015learning}, 
\begin{equation}\label{equ:policy_learning}
    \theta \leftarrow \theta+\alpha \frac{\partial V_{\phi,\theta}}{\partial \theta},
\end{equation}
where $\alpha$ is the learning rate. 

\textbf{Algorithm:} We combine all pieces discussed above and propose the stochastic event driven reinforcement learning (SEDRL) which consists of  three parts: 1. The stochastic agent collects the real data trajectory  from the environment, and train the NHPI by maximizing the likelihood of NHPI. 2. The agent learns the transition and reward model on $\mathbf{\Tilde{h}}(t_i)$ (the state $s(t_i)$ in our algorithm). 3. The agent learns the value function $V_\phi$  and improves the policy using \eqref{equ:policy_learning}. The pseducode of the algorithm is deferred to Appendix \ref{app_section:algorithm}.

\section{Experiment}
\begin{figure*}[t]
\captionsetup[subfloat]{farskip=2pt,captionskip=1pt}
\centering
    \subfloat[8 events (SI )]{\includegraphics[width=.31\textwidth]{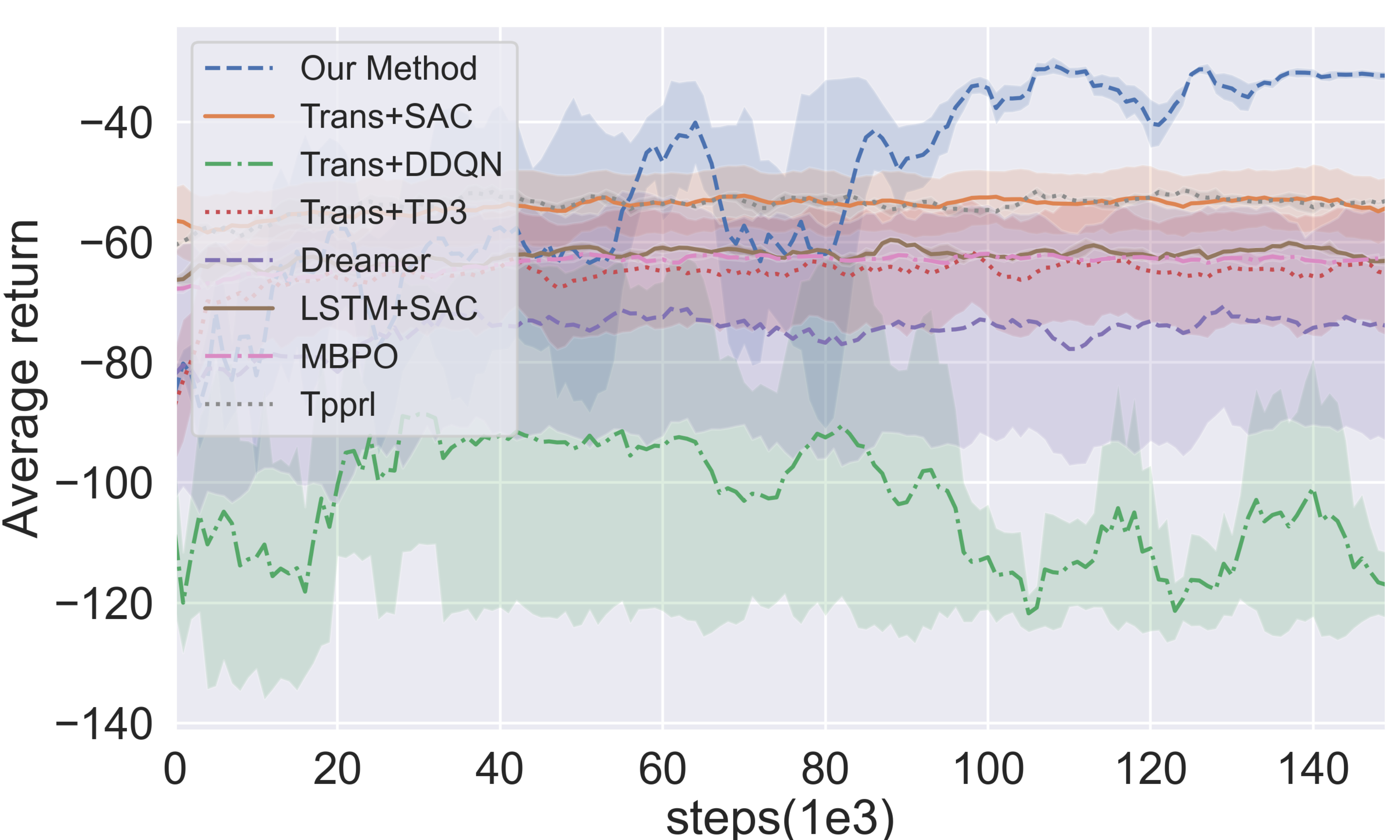}}\hspace{0.01em}
    \subfloat[16 events (SI )]{\includegraphics[width=.31\textwidth]{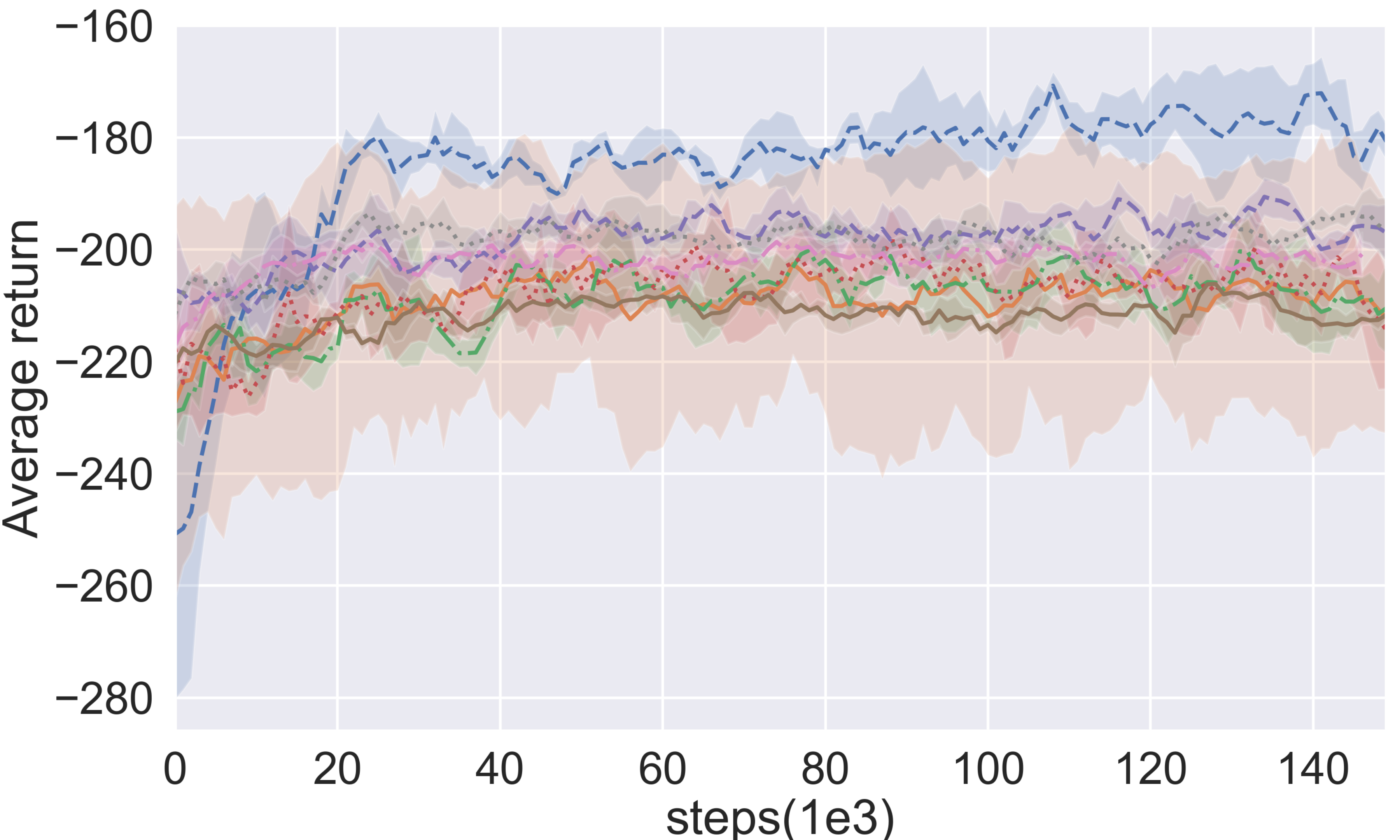}}\hspace{0.01em}
   \subfloat[8 events (USI )]{\includegraphics[width=.31\textwidth]{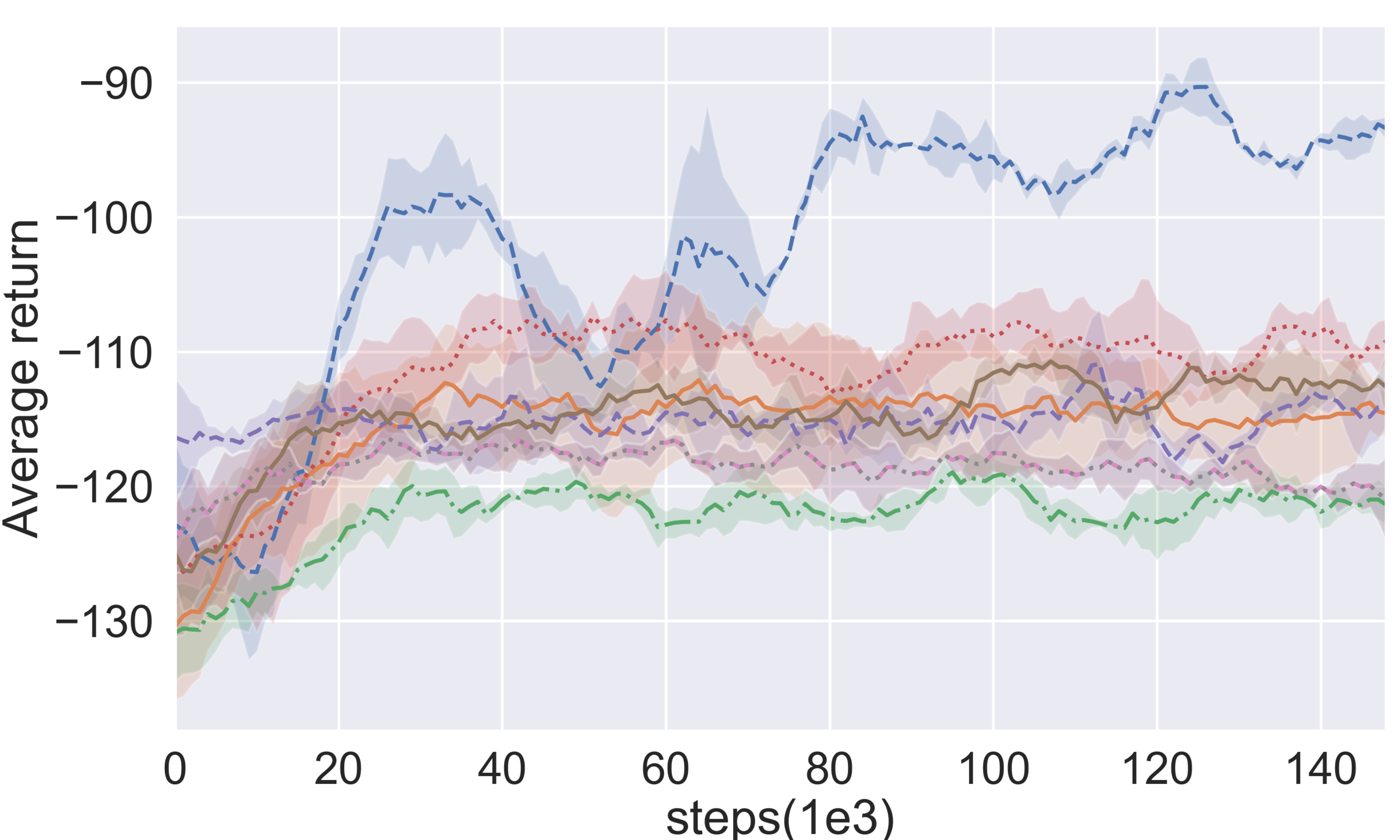}}\hspace{0.01em}
    \subfloat[16 events (USI )]{\includegraphics[width=.31\textwidth]{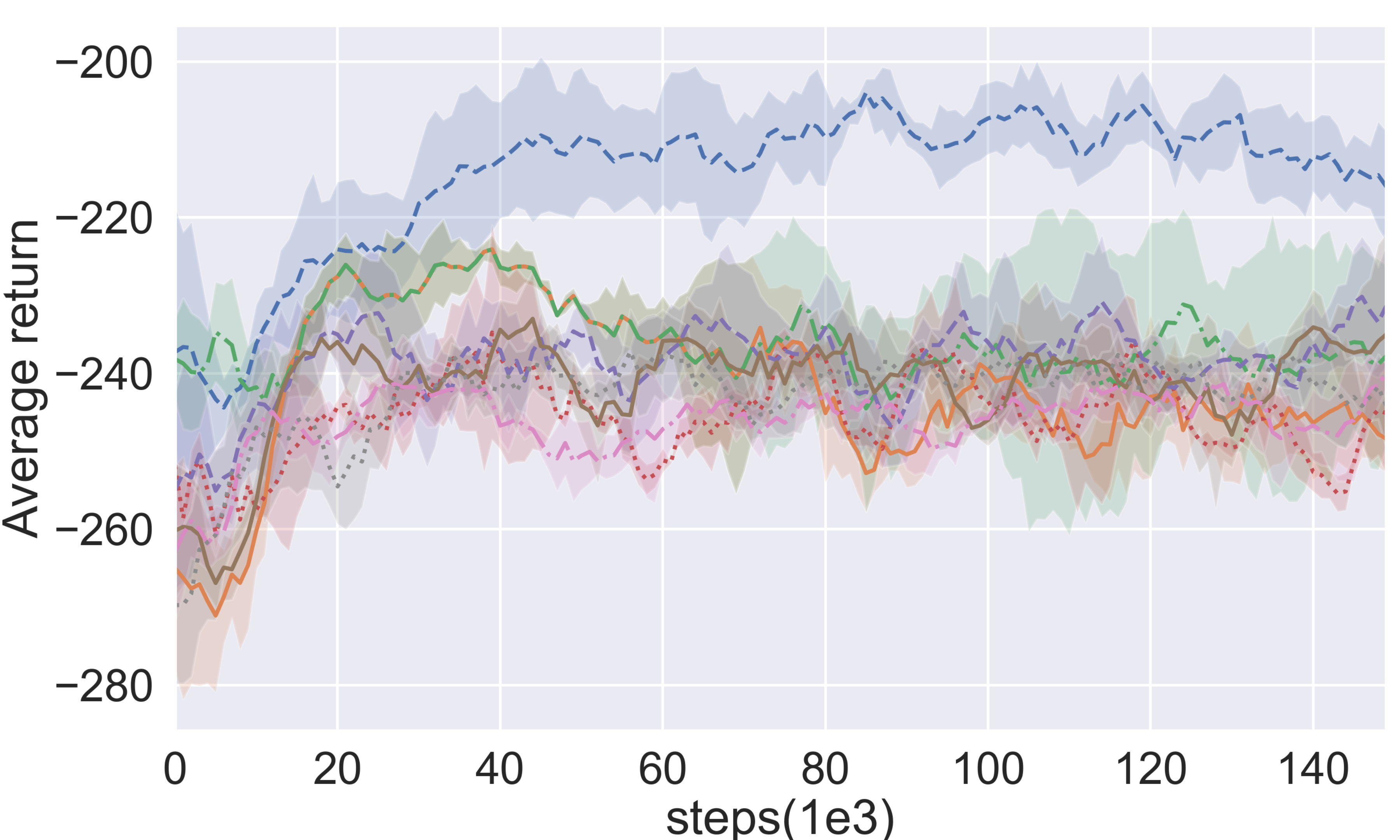}}\hspace{0.01em}
    \subfloat[Smart broadcasting]{\includegraphics[width=.31\textwidth]{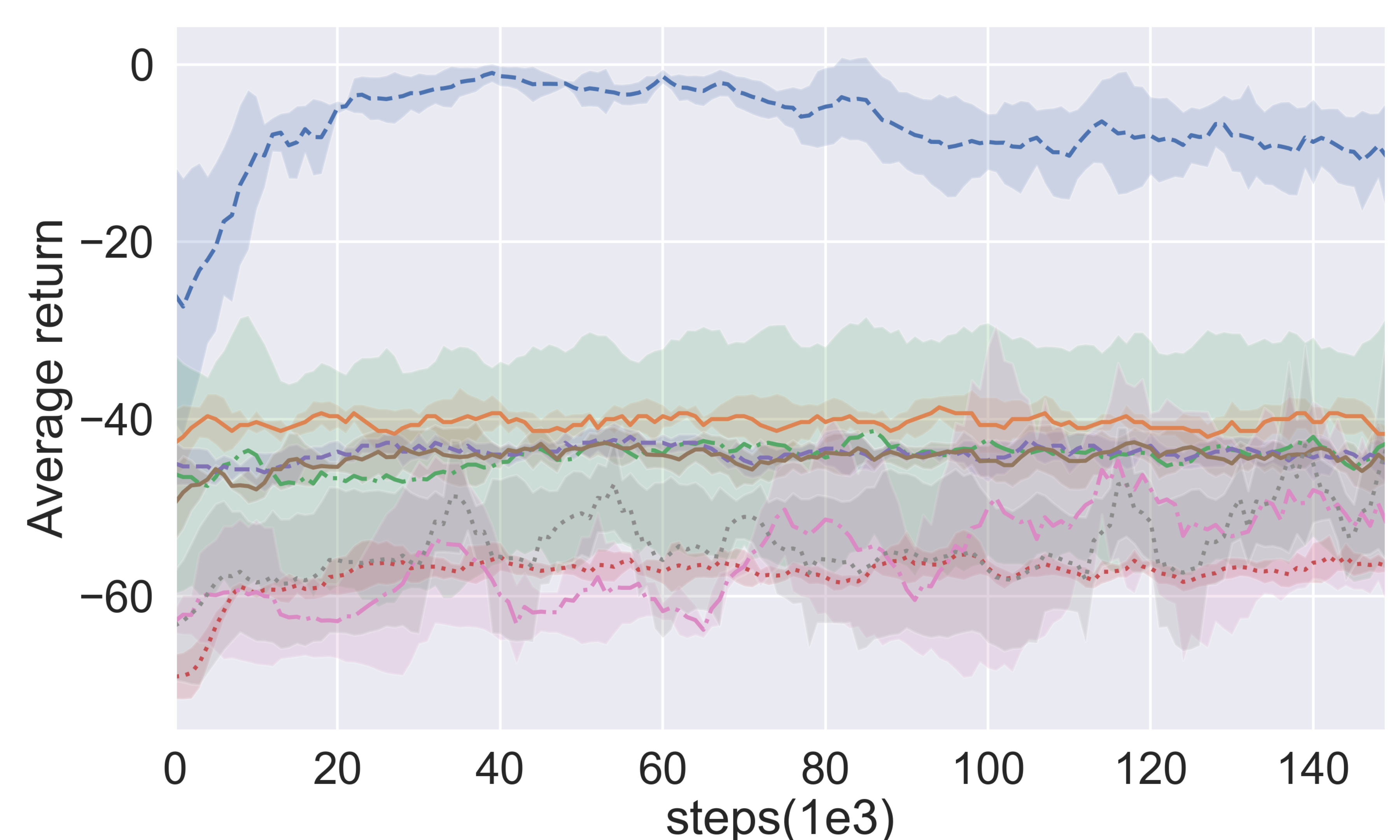}}\hspace{0.01em}
   \subfloat[Improving engagements]{\includegraphics[width=.31\textwidth]{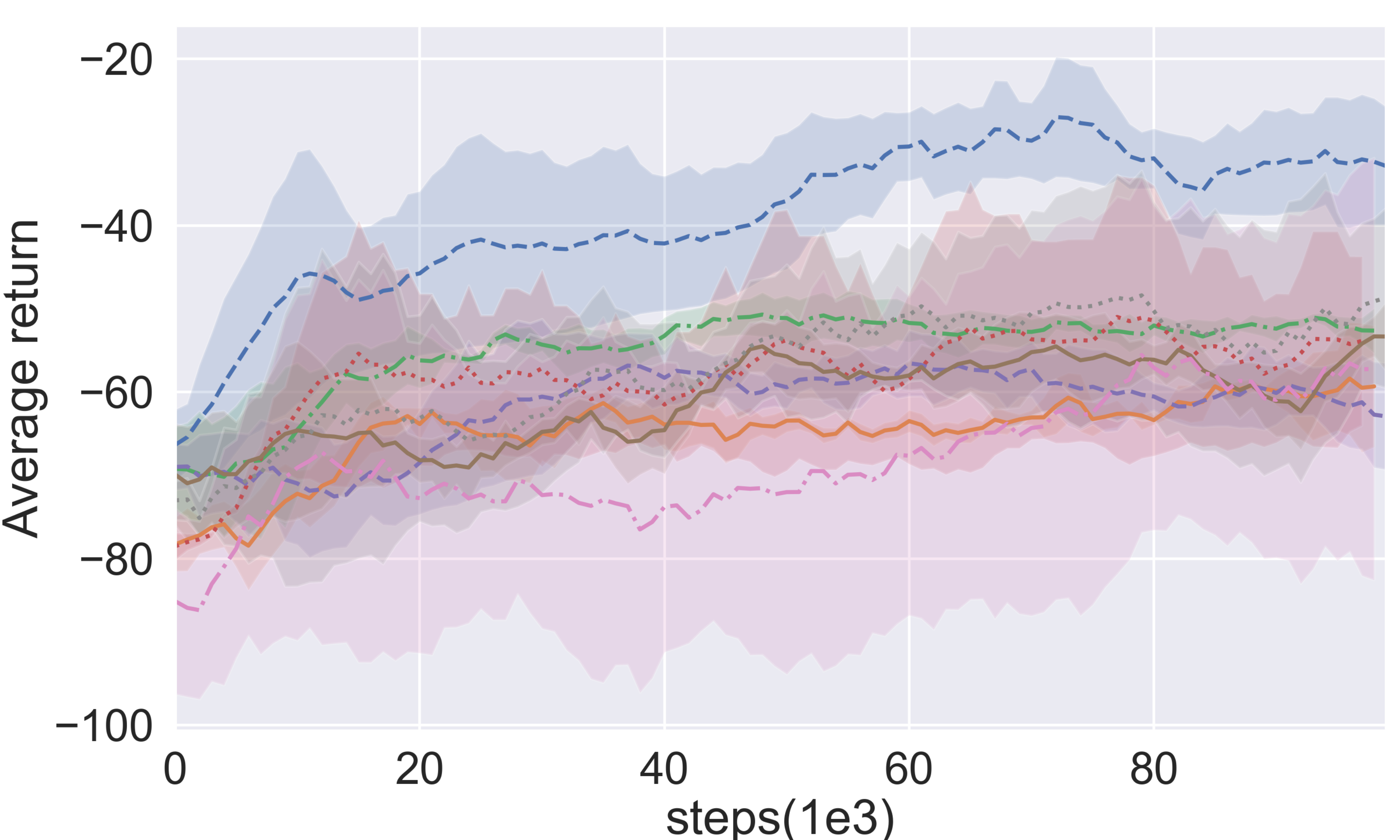}}\hspace{0.01em}
   \caption{ Performance of proposed methods and other baselines. The x-axis is the training step and the y-axis is the cumulative reward. Each experiment is tested on ten trials using various random seeds and initialized parameters. The solid line is the mean of the average return. The shaded region represents the standard deviation.
}\label{fig:experiment}
   \vspace{-5mm}
\end{figure*}

In this section, we evaluate our algorithm SEDRL comprehensively in  the synthetic simulation and experiments with real data such as smart broadcasting and improving the engagement  of the social media platform.  In the real-data experiment, we follow the literature  to build the simulator with real-world data \cite{upadhyay2018deep,zarezade2017steering}, since directly deploying RL agent in the scenarios like social media and healthcare is difficult and may cause safety and ethics issue. We compare our algorithm with multiple state-of-the-art deep RL methods including  model-free algorithms   such as Tpprl \cite{upadhyay2018deep}, SAC \citep{haarnoja2018soft},  TD3 \citep{fujimoto2018addressing}, DDQN \citep{van2016deep} and model-based ones such as Dreamer \citep{hafner2019dream} and MBPO \citep{janner2019trust}. \footnote{There are other related works such as \cite{wang2018stochastic,farajtabar2017fake,zarezade2017steering}, but they either assume the dynamics is known or just can solve simple problem. Therefore we do not report the result here.}  Among them, Tpprl explicitly considers the stochasticity of the time interval $\tau$ and devices a model-free policy gradient algorithm. For the other conventional RL baselines, the problem is formulated as MDP where timestamps are just regular features for the neural networks. We conduct the ablation study on the SMDP formulation in appendix.

 The inputs for all algorithms are the sequences of the observation, i.e., $\{(t_j, k_j ,a_j)\}_{j=1,..,L}$.  We equip  aforementioned  baselines  with LSTM \citep{hochreiter1997long} or Transformer \citep{vaswani2017attention} in the value function network and policy network to encode the observations into state rather than simple feed-forward neural networks as usual, since the environment is highly partially observable and the observation is in the form of sequence \cite{oh2015action}. Notice that in our algorithm the hidden space is obtained by the Transformer structure (multi-head attention+MLP). Therefore the performance gap between our SEDRL and baselines are \emph{only} due to the performance of the algorithm itself. Some details of the baselines are in the following: 
  \textbf{1)} Tpprl:  Tpprl is a model-free RL algorithm tailored for marked temporal point processes \cite{upadhyay2018deep}.
  \textbf{2)} Transformer+SAC:  SAC is a strong baselines in RL and achieves robust performance in almost all test environment \cite{haarnoja2018soft}.  We implement Transformer to encode the event sequence \citep{parisotto2020stabilizing,vaswani2017attention} . Specifically, we stack two multi-head attention layers with head number equal two. Then we feed this embedding as the input to discrete action SAC \citep{haarnoja2018soft,christodoulou2019soft}.  
 \textbf{3)} Transformer+DDQN: This baseline is similar to the Transformer+SAC, but we replace SAC by DDQN.
  \textbf{4)} Transformer+TD3:  We replace SAC by discrete action version of  TD3.    
  \textbf{5)} Dreamer: Dreamer is a model-based algorithm where the deterministic transition model is modeled by LSTM. We replace the CNN in Dreamer by MLP since the input is not image. 
  \textbf{6)} MBPO: MBPO applies the model ensemble in the dynamic model learning to enhance the performance.  We replace the feed-forward neural network in MBPO  by LSTM. 
  \textbf{7)} LSTM+SAC: The sequence encoding is obtained by the LSTM. Other details such as the hyperparameter tunnings in SEDRL and baselines are in Appendix.

\vspace{-3mm}
\subsection{Experimental result}\label{section:synthetic}

\textbf{Synthetic simulation:} 
we simulate the fake news mitigation problem \cite{farajtabar2017fake}. The recent proliferation of malicious fake news in social media has been a source of widespread concern. In this simulation, the platform is the agent and the users are the environment. The social media platform hopes to optimize the performance of real news propagation. Intuitively, it wants to steer the spontaneous user mitigation activities by exposing more valid news.  The action is an event that is selected from some predefined valid news(event) set, which hopefully can distract the attention of users on the rumor.  We create the environment by constructing a simulator of Hawkes process  with exogenous intervention \cite{wang2018stochastic} using Thinning algorithm \citep{ogata1981lewis}. At time step $t_i$, the input of the simulator is the action $a_i$ taken by the agent. The $k$-th action type is encoded as a one-hot vector where $k$-th entry is one. Zero vector means no action taken. The simulator gets the action as an input and generates the event $v_{i+1}\in \{1,...,K\}$ at $t_{i+1}$ ($K$=8 or 16 in our simulation). The parameter such as  $\beta_{ij}$ in the simulator  are some predefined values.  The goal of the platform  is to find the best policy that can control the event (valid and fake news) intensity vector such that it is close to a predefined level $\bm{\lambda}_{target}$. Hence, the reward of our simulation for each time step  $r_t = -\| \bm{\lambda}_{target} - \bm{\lambda}_t \|^2 - 0.1*\text{cost}(a_t)$ provided by the simulator, where $\bm{\lambda}_t$ is the intensity at $t$.  Cost of action is $1$ if the agent inserts an event and otherwise $0$ if we do nothing.

Two intervention settings are considered in our simulation, i.e., synchronized intervention  setting and unsynchronized intervention  setting. We use \emph{SI} and \emph{USI} for short respectively. Notice that  the events occur asynchronously in both settings. The SI setting is straightforward, with the aim  to show that even if in an easy setting, the baselines can not learn a good policy.  Specifically, in the SI setting, the agent can intervene in the sequence at evenly discretized time interval, i.e., at timestamp $\{\Delta t, 2\Delta t,..,i \Delta, ...\} $ (see the illustration in Appendix of experiment setting. This setting is simpler than our original setting (USI) in the section of problem setup, where the intervention can be applied  at $ \{t_1,t_2,...,t_i,..\} $, i.e., at the timestamps of event $i$. The simulation is terminated when time $t$ exceeds the maximum time horizon (e.g., 100 time units.). In general, in this setting, the agent has more chances to intervene with the environment (in our simulation $\Delta< (t_{i+1}-t_i)$). In addition, since time is evenly discretized, this setting may be more friendly to the baselines. Also note that our algorithm (and  baselines) can easily adapt to this setting, since our model can easily accommodate by setting an additional null event occurring at timestamp $i \Delta $. The USI setting is our setting in the  problem setup section, which means the agent only makes decision when a specific event happens along the trajectory (see Figure \ref{fig:problem_setting}), which is more practical in real-world problems.  For instance, the doctor would change the treatment only after the patient's visit.  In our simulation, the agent inserts actions at the next coming clock tick following the occurrence of events.  The simulation is  terminated when the time $t$ exceeds the maximum time horizon. In both settings, we conduct two simulations with K=8 and K=16 events respectively. We use half of these events as our action set (i.e., 4 events and 8 events respectively).  Details of the synthetic simulation are deferred to the appendix of experiment setting.
 All experiments in Figure \ref{fig:experiment} are repeated with ten trials with various random seeds and initial parameters. 
 
Figure \ref{fig:experiment} (a) and (b) demonstrate the results in the SI  setting. In (a), our method SEDRL has the best result followed by Tpprl, Transformer+SAC, LSTM+ SAC, MBPO, Transformer+TD3, Dreamer and then Transformer+DDQN (Trans+Q).  In (b), our SEDRL also achieves the best result. The runner up is Dreamer, while other baselines have similar results around -210. Figure \ref{fig:experiment} (c) and (d) present the results in the USI setting. We can derive similar conclusion as that in the SI setting. Our method outperforms conventional RL algorithm since  we consider the dynamics that can be interrupted by the stochastic event. Comparing with Tpprl, our method explicitly embeds the dynamic model into the Bellman equation of SMDPs and therefore it is much more sample efficient than  the model-free policy gradient algorithm.  

\textbf{Smart broadcasting:} In this problem, the user of the social media wants to decide what to post to elicit the attention from her followers \cite{spasojevic2015post}. The user herself is the agent and she generates action event (post tweets). The environment is formed by her followers and  it generates feedback events if her followers retweet.  Since we cannot make real interventions on the platform of twitter , we follow the conventions in the literature \cite{upadhyay2018deep,zarezade2017steering} to train a simulator with  real-world data. In particular, we use Retweet dataset \citep{zhao2015seismic} to learn a model of follower. The Retweets dataset contains sequences of tweets, where each sequence contains origin tweets (i.e., some user initiates a tweet), and some follow-up tweets. We group the post and retweet into three categories event (i.e., $K$=3) following the work in   \citep{zuo2020transformer}.  Then follower's behavior is learned by a neural temporal point process \citep{zuo2020transformer} with exogenous excitation (i.e., the origin tweets) \cite{ding2020traffic}. The users should expose the predefined events (action set are three kinds of post) to her followers to elicit attentions and then observe the feedback event of the followers. Each kind of the feedback event (retweet) and action event (post) corresponds to a predefined reward and cost respectively( appendix G), where the aim of action cost is to discourage too much intervention over the environment. The goal of the agent is to maximize the total reward minus the cost over the simulation time [0,T]. Given the feedback model of the follower (generated by the simulator), we can do simulations through the interactions over the environment.  Some details of the experiment setting are deferred to the appendix . In Figure \ref{fig:experiment} (e), we demonstrate the performance of all algorithm. Our method reaches asymptotic result after only 20k training steps and achieves $-10$ long-term reward, while the performances of baselines are below $-40$.

\textbf{Improving the Engagement:} The platforms of e-commerce or social media sometimes award the users to promote engagement in the community. In this problem, we would design a policy to help the platform to hand out awards. Take Stack-Overflow for instance, where users can use post to ask questions or seek advises in various technical areas while the other users can answer the questions freely and the answer will be rated or awarded by other views. Under our problem setting, the agent is the platform, while the environment consist of the users in the website.  The action event  correspond to badge types, e.g., Nice Question, Good Answer, etc, and the event times correspond to the award times. The platform  can encourages user engagement by giving certain types of badges (action) to the user after he receives some other badge.  For example, when a user receives ``Good Answer'' badge, the agent  awards a ``Great Answer'' or ``Guru'' badge (as an action)  to encourage his/her engagement in the website. Then the platform can observe the feedback event from  the users. For instance, The user may answer  questions more frequently with higher quality (e.g., more useful answers).  Each feedback event and action event have a corresponding predefined reward or cost. We sum  all of them over a certain period to reflect the engagement rate of the platform. The agent should learn a control policy to maximize the long-term reward by manipulating the action events. Since we can not deploy the policy in the Stack-Overflow, we use the data gathered from Stack-Overflow \cite{snapnets} to learn a feedback model of the users.  The data is collected in a two-year period, and we treat each user’s reward history as a sequence following the work \citep{zuo2020transformer}. We pick eleven types of the event in the dataset as the exogenous excitation while the other half as the endogenous one.    Therefore, the size of the action set is eleven and $K=22$ (See more details in appendix \ref{app_section:exp_setting}).  We can learn a user feedback model using neural temporal process with exogenous excitation \cite{ding2020traffic}. Using this simulator (feedback model) trained from the real data set, we can do experiments through the interactions with the environment. We report results in Figure \ref{fig:experiment} (f).   Our method could achieve $-35$ cumulative reward after 100k step training, while the rewards of baselines are below $-50$.
% \vspace{-2mm}
% \subsection{Ablation study}\label{section:ablation_study}
% \vspace{-2mm}
% In the ablation study, we adapt the SAC into its SMDP version by modifing its $V$ and $Q$ function according to bellman equation in SMDPs. Note that this modified model itself is novel. We show that even we take into account SMDPs, the performance gap between SAC and our algorithm is still quite large.  We also test the performance of SAC algorithm only using the multilayer perceptrons structure, even though it is well-expected that such structure does not work well.  We discuss more details and demonstrate the results in Appendix \ref{app:ablation}.
% \vspace{-3mm}
% \section{Conclusion and future work}
% \vspace{-3mm}
% Inspired by the real world problem where people interact with a complex environment by means of asynchronous stochastic discrete events in continuous time, we proposed a novel model-based RL algorithm. It captures the dynamics with jumps in the sequence of stochastic event via Hawkes process and learns a sample-efficient policy to maximize the long-term reward. We demonstrate its superiority in synthetic and real data experiment. One possible limitation of our work is that we use the dynamic model to generate imaginary data to train the agent, which may cause the model bias in RL. In the future work, we would combine model-based and model-free method together to mitigate such issue.

% Use \bibliography{yourbibfile} instead or the References section will not appear in your paper
% \nobibliography{aaai23}
\bibliography{event_driven}

\newpage
\appendix
\onecolumn

\section{Organization of Appendix}
\begin{itemize}
    \item SDE reformulation and connection to  World Model and Neural ODEs
    \item Parameterization of Neural Hawkes Process with Intervention
    \item Log-Likelihood of Neural Hawkes Process with Intervention
    \item Estimation of the Integration in the log-likelihood
    \item Algorithm details
    \item Transition,Value Function and Policy Learning
    \item Experiment Setting
    \item Settings of baselines and hyperparameters
    \item Computational resource
    \item Ablation Study
    \item Limitations and Future work
\end{itemize}

\section{SDE reformulation and connection to  World Model and Neural ODEs} \label{section:SDE_reformulation}

To begin with, we link Hawkes process to the SDEs with jumps.  Then we discuss why our  NHPI-based work is different from the conventional model-based RL. First, notice that Hawkes process can be reformulated as an SDE with jumps \citep{wang2018stochastic}, in which
\begin{equation}\label{equ:sde}
   d\lambda_k(t)=  \zeta\big(\mu_k-\lambda_k(t)\big)dt + \sum_{j=1}^{K} \beta_{jk}  dN_j(t) ,
\end{equation}
where $N_j(t)$ is the  counting process for event $j$.

We follow the proof in \citep{wang2018stochastic} and present here for completeness. To begin with, we define the convolution operator $*$ as $f(t)*g(t) = \int_{0}^t f(t-s)g(s)ds$.

Recall that we have Hawkes model  as follows
\begin{equation}\label{app_equ:intensity_hawkes_action}
      \lambda_k(t) = \mu_k +\sum_{h: t_h<t}\beta_{k_h,k} \exp (-\zeta(t-t_h)), k=1,...,K
\end{equation}

We can rewrite the term $\sum_{h: t_h<t}\beta_{k_h,k} \exp (-\zeta(t-t_h))$, i.e., effects of the all past events on the current event with type $k$ as 

\begin{equation}\label{app_equ:convolution}
     \sum_{j=1}^{K}\beta_{jk}\exp(-\zeta t)*dN_j(t).
\end{equation}

Using the basic knowledge of the calculus, we know $d(f*g) = f(0)g+ g*df$.

Take differential operator $d$ at the both sides of \eqref{app_equ:intensity_hawkes_action} and plug in above result and note that $\mu_k$ does not depend on $t$, we have

\begin{equation}
\begin{split}
     d\lambda_k(t) &= d\mu_k + d \big(\sum_{j=1}^{K}\beta_{jk}\exp(-\zeta t)*dN_j(t)\big)\\
     & = \sum_{j=1}^{K} \beta_{jk}  dN_j(t)- \underbrace{\sum_{j=1}^{K} dN_j(t)* \beta_{jk} \zeta exp(-\zeta t)dt}_{(a)}\\
     &\overset{(1)}{=}  -\zeta \big(\lambda_k(t)-\mu_k\big)dt+ \sum_{j=1}^{K} \beta_{jk}  dN_j(t),\\
\end{split}
\end{equation}
where (1) holds because we use the fact that the term $(a) = \zeta(\lambda_k(t)-\mu_k)dt$ from \eqref{app_equ:intensity_hawkes_action} and \eqref{app_equ:convolution}.

In  equation \ref{equ:sde}, $\zeta\big(\mu_k -\lambda _k (t)\big)dt$ is a drift term. $-\lambda_k(t) dt$ is a force in the negative direction, which stabilizes the process and pushes it to the base intensity  $\mu_k$. The endogenous jump term is $\sum_{j,k}\beta_{jk} dN_j(t)$ which captures the influence of abrupt events. Recall that $N_j(t)$ is the counting process of event of type $j$. $dN_j(t) = 1$ when the event of type $j$ happens\footnote{ Under the assumption that two events do not happen simultaneously, which is a common assumption in temporal point process.} while $dN_j(t)=0$ at other times. Hence this term reflects how the events that occurred before $t$ affect  event of type $k$ in the future. One direct way to incorporate the influence of action is 
\begin{equation}\label{equ:SDE_reformulation}
    d\lambda_k(t)= a_k(t)dt+  \zeta\big(\mu_k-\lambda_k(t)\big)dt + \sum_{j=1}^{K} \beta_{jk} dN_j(t)
\end{equation}

If we replace the endogenous \emph{jump} terms by some \emph{continuous} terms and then  approximate  $\mu_k, \beta_{jk}, \zeta$ by neural networks such as RNN or its variants with input event streams, Equation \eqref{equ:SDE_reformulation} reduces to the neural ODE \citep{chen2018neural,du2020model,yildiz2021continuous}. Therefore both ODEs approaches and our method allow the environment to be discontinuously updated by exogenous actions (of RL agents); our formulation allows the environment dynamics to have endogenous jumps, which can not be handled by ODEs. By solving this ODE with \emph{even} time interval, we would get the transition function with a recurrent form  $ \bm{\lambda}(t+1) = f( \{\bm{\lambda}(\tau),\mathbf{a(\tau)}\}_{\tau=1...t})$,  where the bold $\bm{\lambda}$ is a vector and the $k$-th element of it is $\lambda_k$. In the partially observable case, we may get a transition function on hidden vector $h$ of RNN. Such recurrent transition model can be thought of as the hidden space of the world model\citep{ha2018world}. Its variants may include \cite{hafner2019dream} when combining with VAE (if the inputs are images) and many other conventional wisdom of model-based RL can apply to this model. However, there are several downsides if we follow this approach: 1) World model neglects the nature of the dynamics in the discrete event stream, i.e., the jump term that captures the abrupt events in the SDEs. Besides it can not solve the problem with irregular time intervals, since its dynamic model (e.g., RNN) discretizes time evenly.  2) The variants of Neural ODE, such as latent ODE \citep{latentode2019} and Jump SDE \cite{jia2019neural} have the capabilities of modeling discrete event streams. However, if we directly apply the Neural ODE approach, we need expensive numerical solvers with additional issues of convergence stability and data-efficiency \citep{chen2020neural}.

\section{Parameterization of Neural Hawkes Process with Intervention}\label{app_section:parameter_NIHP}

 As shown in the Figure \ref{fig:NIHP}, we encode the event types, timestamps and actions into dense embedding. $\mathbf{U}\in \mathbb{R}^{M\times K}$ is an embedding matrix for the event types, where the $k$-th column is an $M$-dimensional embedding for event type $k$. Let $\mathbf{k}_i$ be the onehot encoding of the event with type $k_i$, then the embedding of event at timestamp $t_i$ is $\mathbf{U}\mathbf{k}_i$.

We then encode the timestamps following the work in \citep{zuo2020transformer}  to cope with the challenge of the irregular time interval . \emph{Timestamp encoding} is 
\begin{equation}\label{app_equ:time_encoding}
  [\mathbf{z}(t_i)]_m =
  \begin{cases}
 cos(\omega_m t_i +\omega_m i)      &  \text{ m is even}\\
    sin(\omega_m t_i +\omega_m i)   & \text{ m is odd},
  \end{cases}
\end{equation}
where trigonometric functions are used  to define a temporal encoding for timestamps $t_i$,  $\omega_m$ is a predefined frequency, and the vector $\mathbf{z}(t_i) \in \mathbb{R}^M$. Other temporal encoding methods can also be
applied, such as the relative position representation model \citep{shaw2018self}. The action $a_i$ pass an MLP to get an $N$-dimensional \emph{action embedding} $\mathbf{a}_e(t_i)$.

Embedding of the event sequence $Es=\{ (t_i,k_i,a_i)\}_{i=1}^{L}$ is then specified by 
$$\mathbf{X}=[(\mathbf{UY}+\mathbf{Z})^T,\mathbf{A}^T],$$
where $\mathbf{Y}=[\mathbf{k}_1, \mathbf{k}_2,...,\mathbf{k}_L]\in \mathbb{R}^{K\times L}$ is the collection of event type embedding, $\mathbf{Z}=[\mathbf{z}(t_1),\mathbf{z}(t_2)...,\mathbf{z}(t_L)]\in \mathbb{R}^{M\times L}$ is the collection of timestamp encoding, while $ \mathbf{A}=[\mathbf{a_e}(t_1),\mathbf{a_e}(t_2)...,\mathbf{a_e}(t_L) ] \in \mathbb{R}^{N\times L} $ is the collection of action embedding. Then the embedding $\mathbf{X}$ passes through a self-attention module.
\begin{equation}\label{equ:attention}
    \mathbf{\Lambda} = softmax(\frac{\mathbf{Q} \mathbf{K}^T}{\sqrt{M_K}})\mathbf{V},
\end{equation}
where $\mathbf{Q}=\mathbf{XW}^Q$, $\mathbf{K}=\mathbf{XW}^K, \mathbf{V}=\mathbf{XW}^V$ are the query, key and value matri, respectively . $\mathbf{W}^Q,\mathbf{W}^{K}\in \mathbb{R}^{(M+N)\times M_{K}}$, $\mathbf{W}^V\in \mathbb{R}^{(M+N)\times M_V}$ are learned weight matrix. To avoid observing the furture events, the attention is equipped with masks, specifically when computing the attention $\mathbf{S}(j,:)$, we mask all the future positions.

Then we feed the self-attention $\mathbf{\Lambda}$ into a feed-forward neural network to generate the hidden representation $\mathbf{H}$:
$$ \mathbf{H} = Relu(\mathbf{\Lambda W}_1+\mathbf{b}_1)\mathbf{W}_2+\mathbf{b}_2.$$

The hidden representation at time step $t_i$ is the $i$-th column of $\mathbf{H}$, i.e.,  $\mathbf{h}(t_i) = \mathbf{H}(i,:)$.

\section{Log-Likelihood of Neural Hawkes Process with Intervention}\label{app_section:likelihood}
In this section, we provide the log likelihood for the neural Hawkes process with intervention. The proof mainly follows the conditional intensity function approach in \citep{rasmussen2018lecture}. 

We start with the unmarked case, i.e., there is just one type event. Then we generalize to the case with $K$ types (marked case). 
Let $f(t_{n+1}|\mathcal{H}_{t_n})$ be the conditional density function of the time of the next event time  $t_{n+1}$ given the history  $\{ (t_1,a_1),...,(t_n,a_n) \}$ with $t_n<t$. Here we abuse the notation history $\mathcal{H}_{t_n}$ a little here to include the actions.  Note that here we only want to calculate the conditional  probability of event given the action in the sequence.  Let $F(t|\mathcal{H}_{t_n})$ be the corresponding cumulative distribution function.

Our first claim is that 
\begin{equation}\label{app_equ:conditional_intensity}
    \lambda(t) = \frac{f(t|\mathcal{H}_{t_n})}{1-F(t|\mathcal{H}_{t_n})}.
\end{equation}

We prove this claim in the following.
\begin{equation}\label{app_equ:derivation}
\begin{split}
\frac{f(t|\mathcal{H}_{t_n})dt}{1-F(t|\mathcal{H}_{t_n})} = &\frac{\mathbb{P}(t_{n+1}\in [t,t+dt]|\mathcal{H}_{t_n})    }{\mathbb{P} (t_{n+1}\notin(t_n,t)|\mathcal{H}_{t_n} )}\\
=&\frac{\mathbb{P}(t_{n+1}\in [t,t+dt],t_{n+1} \notin (t_n,t)|\mathcal{H}_{t_n}) }{\mathbb{P}(t_{n+1}\notin(t_n,t)|\mathcal{H}_{t_n} )  }\\
=&\mathbb{P}(t_{n+1}\in [t,t+dt]|t_{n+1}\notin (t_n,t),\mathcal{H}_{t_n} )\\
=&\mathbb{P}(t_{n+1}\in [t+dt]|\mathcal{H}_{t-})\\
=& \mathbb{E}[N([t,t+dt])|\mathcal{H}_{t-}]\\
=&\lambda(t)dt,
\end{split}
\end{equation}
where $\mathcal{H}_{t-}$ is the knowledge of time of all events up to but not including time $t$, and $N([t,t+dt])$ denotes the number of points falling in an interval $[t,t+dt]$.

Using \eqref{app_equ:conditional_intensity}, we have 

\begin{equation}
\lambda(t) = \frac{\frac{d}{dt} F(t|\mathcal{H}_{t_n}) }{1-F(t|\mathcal{H}_{t_n})} = -\frac{d}{dt}\log(1-F(t|\mathcal{H}_{t_n})).
\end{equation}

Therefore

\begin{equation}\label{app_equ:int_labmda}
    \int_{t_n}^{t}\lambda(s)ds = -(\log(1-F(t|\mathcal{H}_{t_n})))+\log(1-F(t_n|\mathcal{H}_{t_n})) = -\log(1-F(t|\mathcal{H}_{t_n})),
\end{equation}
where we use the fact that $F(t_n|\mathcal{H}_{t_n})=0$.

Using \eqref{app_equ:int_labmda}, we obtain 

\begin{equation}\label{app_equ:cumulative_distribution}
    1-F(t|\mathcal{H}_{t_n}) = \exp(-\int_{t_n}^t\lambda(s)ds)
\end{equation}
and 
\begin{equation}\label{app_equ:probability_density}
    f(t|\mathcal{H}_{t_n})=\lambda(t)\exp(-\int_{t_n}^t \lambda(s)ds).
\end{equation}

For the marked case with total $K$ types, we can specify the distribution of the event $k$ associated with the point $t$ by its conditional density function $f^*(k|t)=f(k|t,\mathcal{H}_{t-})$, which specifies the distribution of the event $k$ given $t$ and the history $\mathcal{H}_{t-}$ including the information of timestamp, action and event type. Note that now the history also includes event type. Remark that $f^{*}(k|t) = f(k|t,\mathcal{H}_{t_n})$ if $t_n$ is the last point before $t$, since the additional condition that the next point is located at $t$ means that the history $\mathcal{H}_{t-}$ and $\mathcal{H}_{t_n}$ contain the same information (Recall that the history $\mathcal{H}_{t-}$ does not include time $t$, while $H_{t_n}$ includes timestamp $t_n$.)
%You may use another notation $\mathcal{H}_{t-}$ to replace $\mathcal{H}_t$ to ease the understanding)

The conditional intensity function for the marked case is thus
$$ \lambda_k(t) = \lambda(t)f^*(k|t),$$
where $\lambda(t)$ is called ground intensity and is exactly the conditional intensity function for the unmarked case.

Using the exactly same argument in \eqref{app_equ:derivation}, we have

\begin{equation}\label{app_equ:marked_intensity}
    \lambda_k(t) = \frac{f(t|\mathcal{H}_{t_n})f^*(k|t) }{1-F(t|\mathcal{H}_{t_n})} = \frac{f(t,k|\mathcal{H}_{t_n})}{1-F(t|\mathcal{H}_{t_n})}
\end{equation}

The likelihood function is the joint densithy function of all the points $\{(t_1,k_1),...,(t_n,k_n)\}, t_n< T$,  given the action $\{a_1,...,a_n\}$.
$$ L =  f(t_1|\mathcal{H}_{t_0}) f(k_1|t_1,\mathcal{H}_{t_0}) \cdots f(t_n|\mathcal{H}_{t_{n-1}})f(k_n|t_n,\mathcal{H}_{t_{n-1}})(1-F(T|\mathcal{H}_{t_n}))$$
where the term $(1-F(T|\mathcal{H}_{t_n})) $ accounts for the fact that $t_{n+1}$ happends after $T$.
\begin{equation}
\begin{split}
      L = &  \bigg(\prod_{i=1}^{n} f(t_i|\mathcal{H}_{t_{i-1}})\bigg)\bigg( \prod_{i=1}^{n} f(k_i|t_i,\mathcal{H}_{t_{i-1}})\bigg)(1-F(T|\mathcal{H}_{t_n}))\\
      \overset{(1)}{=}& \bigg( \prod_{i=1}^{n} \lambda(t_i)\exp(-\int_{t_i-1}^{t_i} \lambda(s)ds )   \bigg) \bigg( \prod_{i=1}^{n} f(k_i|t_i,\mathcal{H}_{t_{i-1}})\bigg)\exp \big(-\int_{t_n}^T \lambda(s)ds\big)\\
      =&\bigg(\prod_{i=1}^n \lambda(t_i) \bigg)\bigg( \prod_{i=1}^{n} f(k_i|t_i,\mathcal{H}_{t_{i-1}})\bigg)  \exp\big(-\int_{0}^T \lambda(s)ds\big)\\
      =& \bigg(\prod_{i=1}^{n} \lambda(t_i)f(k_i|t_i,\mathcal{H}_{t_{i-1}})  \bigg)\exp \big(-\int_{0}^T \lambda(s)ds\big)\\
      =&\bigg( \prod_{i=1}^{n}\lambda_k(t_i)\bigg)\exp \big(-\int_{0}^T \lambda(s)ds\big),
\end{split}
\end{equation}
where $(1)$ holds using \eqref{app_equ:cumulative_distribution} and \eqref{app_equ:probability_density}, the last equality holds because $f(k_i|t_i,\mathcal{H}_{t_{i-1}})= f(k_i|t_i,\mathcal{H}_{t_{i}-})$.

\section{Estimation of the Integration}\label{app_section:estimation_integration}

The only difficulty to estimate the gradient ofin the likelihood of NHPI is the integration term $\int_{0}^{T} \lambda (t) dt$. We can sample $ t\sim Unif(0,T)$. The term $T\lambda(t)$ is the unbiased estimation since its expectation is $\int_{0}^{T} \lambda (t) dt$. In practice, we may average over several samples to reduce the variance.

\section{Algorithm details} \label{app_section:algorithm}

We present the pseudo-code in Algorithm \ref{alg:example}

\begin{algorithm}[H]
   \caption{SEDRL}
   \label{alg:example}
\begin{algorithmic}
   \STATE {\bfseries Input:} Replay buffer $\mathcal{D}$, replay buffer $\mathcal{D}_{tr}$ to store the event trajectory , policy $\pi_\theta$, value function $V_\phi$, dynamic model $g$, reward model $\kappa$
   \STATE{\bfseries Train NHPI: } Using initial policy to collect $M$ trajectories of event streams $\{ \{(t^m_i,k^m_i,a^m_i)\}_{i=1}^{L}\}_{m=1}^M$  and store them in $\mathcal{D}_{tr}$. Maximize the likelihood of NHPI.
   \FOR{each step }
   \STATE {\bfseries 1. Interact with environment} \\
    Obtain state $s(t_i)$ using NHPI and sample the action $a(t_i)$ from the  stochastic policy $\pi_\theta(s(t_i))$. Impose the action on the environment and obtain $\tau_i$,$r(t_i)$, $s(t_{i+1})$. Append tuple $(s(t_i), a(t_i), \tau_i, r(t_i), s(t_{i+1}))$ into replay buffer $\mathcal{D}$.
   \STATE {\bfseries 2. Learn transition and reward model }\\ Sample data from $\mathcal{D}$, update neural network $g$ and $\kappa$ using \eqref{equ:transition_learning}, \eqref{equ:reward_learning}.
   \STATE{\bfseries 3. Value function learning}: Update the value function with \eqref{equ:value_learning}. There are two options to estimate the target function $\hat{V}$ (see details in \ref{app_section:value_function_learning} ):\\
   option 1: sample $s(t_i)$ from $\mathcal{D}$ and use g and NIPH to generate $s(t_{i+1})$ and $\tau_i$ (thinning algorithm).\\ 
   option 2: sample $s(t_i)$ from $\mathcal{D}$. Update $Q$ and $V$ function using \eqref{app_equ:error_q_function},\eqref{app_equ:error_v_function}.
   
   \STATE{\bfseries 4. Policy improvement}
   Update policy with \eqref{equ:policy_learning} (see details in \ref{app_section:policy_learning})
   \STATE{\bfseries Update NHPI:}
   \IF{ a trajectory finishes}
   \STATE Append the new event trajectory to $\mathcal{D}_{tr}$. Sample trajectory in $\mathcal{D}_{tr}$ and maximize the likelihood of NHPI.
   \ENDIF
   \ENDFOR
\end{algorithmic}
\end{algorithm}

\section{ Transition,Value Function and Policy Learning in Algorithm \ref{alg:example}}\label{app_section:value_function}

In this section, we present the detail of the learning of transition, value function from the replay buffer  and policy learning from the imagination, which are deferred from the section \ref{section:planning_latent} due to the limitation of the space.

To begin with, we briefly introduce the thinning algorithm \cite{ogata1981lewis}, which can simulate the time interval and event given the intensity $\lambda_k$.

\subsection{Thinning algorithm:}
 Thinning algorithm is a common algorithm to simulate the temporal point process including  Hawkes process \cite{ogata1981lewis,mei2017neural}. Here we just give a sketch and the readers just need to remember that we can use it to sample the next timestamp $t_{i+1}$ given we are at timestamp $t_i$. Please refer to \cite{rizoiu2017tutorial} for details. Generally speaking, it simulates the inter-arrival time $\tau_i$. For instance, the inter-arrival times in a homogeneous Poisson process with intensity $\lambda$ follow an exponential distribution.  Therefore we can use inverse transform sampling technique. Specifically, we can sample $u$ from a uniform distribution $U(0,1)$, then construct inter-arrival time $\tau= \frac{-ln u}{\lambda}$. The event type is determined by the conditional probability $\lambda_i/\lambda$. For non-homogeneous Poisson process, e.g. Hawkes process, it needs an additional rejected sampling step. Please refer to section 1.4 in  \cite{rizoiu2017tutorial} for details.

\subsection{Transition learning:}
Recall that we already have the state $s_t$ through the NHPI model. When learning the transition and reward model,  i.e., minimizing the error  \eqref{equ:transition_learning} and \eqref{equ:reward_learning}, we fixed the parameters of the network of NHPI. It says the gradient of $g$ and $\kappa$ does not back propagate the network of NHPI.
 
\subsection{Value function learning:}\label{app_section:value_function_learning}
We need the value function to do the policy improvement in the stochastic value gradient step. 
Recall that  Bellman  equation in the SMDPs is 
\begin{equation}\label{app_equ:bellman_equation}
V(s)= \sum_{s',a} \pi(a|s)P(s'|a,s) [\int_{0}^{\infty} \int_{0}^{t}e^{-\rho\tau}r(\tau)d\tau dF(t|s,a,s')
     +   \int_{0}^{\infty} e^{-\rho t} V(s')dF(t|s,a,s')].
\end{equation}

We can do the sampling $s'\sim P(\cdot|a,s)$, $a\sim \pi(\cdot|s)$ in the RHS and obtain

\begin{equation}\label{app_equ:median_sampled_term}
    \int_{0}^{\infty} \int_{0}^{t}e^{-\rho\tau}r(\tau)d\tau dF(t|s,a,s')
     +  \int_{0}^{\infty} e^{-\rho t} V(s')dF(t|s,a,s').
\end{equation}  

The next step is to do the sampling over the time interval. In particular, we sample $t$ according to $dF(t|s,a,s')$.  Note that \eqref{app_equ:median_sampled_term} is in the integration form $\int_{0}^{\infty} (\cdot) dF(t|s,a,s) $ w.r.t $dF(t|s,a,s')$, i.e., $\mathbb{E}_{t\sim dF(\cdot|s,a,s')} (\cdot)$. Therefore we can get a sample $t \sim dF(\cdot|s,a,s')$, which is 
\begin{equation}
    \begin{aligned}
    \int_{0}^{t}e^{-\rho\tau}r(\tau)d\tau
     +   e^{-\rho t} V(s').\\
    \end{aligned}
\end{equation}

% Then we can sample $s'$ w.r.t. $P(s'|a,s)$ and obtain the sampled version of $\int_{0}^{t}e^{-\rho \tau } r(\tau) +e^{-\rho t} V(s')$.

In general, $\int_0^{t} e^{-\rho \tau }r(\tau) $ is not easy to estimated for which we assume $r(\tau)=r(0)$, i.e., reward does not change over time between the transition $s$ to $s'$.  In other words, we assume $r(t) =r(t_i)$ for $t_i\leq t < t_{i+1}$. In practice, this approximation works well.

Therefore the final sampled version for the Bellman equation is
$$ \hat{V}(s):= \frac{1-e^{-\rho t}}{\rho} r(0)+e^{-\rho t} V(s').  $$

We replace $s$ by $s(t_i)$ and $s'$ by $s(t_{i+1})$. Let $\tau_i= t_{i+1}-t_{i}$, we have
\begin{equation}\label{app_equ:target}
    \hat{V}(s(t_i)):= \frac{1-e^{-\rho \tau_i}}{\rho} r(t_i)+e^{-\rho \tau_i} V(s(t_{i+1})).
\end{equation}

Note that, in our method, we need a value function $V_\phi$ of \emph{current} policy $\pi_{\theta}$. There are generally two options to estimate that.

The \emph{first} one is to fully use the imaginary data generated from the NHPI, transition model and reward model. In particular, given a state $s(t_i)$(that can be obtained from NHPI and event trajectory in $D_{tr}$), we sample an action from current policy $\pi_\theta(s(t_i))$ and obtain next state $s(t_{i+1})$ using $g$. Reward $r(t_i)$ can be obtained from learned reward model.  Note that using NHPI and thinning algorithm, we can simulate next timestamp $t_{i+1}$.  Therefore we have all element needed in \eqref{app_equ:target} and then minimize the error $\|V_\phi(s(t_i))-\hat{V}_\phi(s(t_i))\|_2^2$, where $s(t_i)$ is obtain from NHPI and trajectory in replay buffer $D_{tr}$. 

The \emph{second} one is to use real data (we use this approach in our implementation). In this approach, we  need to introduce  $Q$ function, where $Q$ is approximated by a neural network $Q_\psi$. In particular, we minimize  error of $V$ function 

\begin{equation}\label{app_equ:error_v_function}
    \mathbb{E}_{s(t_i)\sim D}[V_{\phi}(s(t_i))-\mathbb{E}_{a\sim \pi_{\theta}} Q_\psi(s(t_i),a) ]^2.
\end{equation}
If the action space is discrete, we can directly calculate $\mathbb{E}_{a\sim \pi_\theta}Q_\psi(s(t_i),a)$ and get the value function $V$.

The error of $Q$ function is
\begin{equation}\label{app_equ:error_q_function}
    \mathbb{E}_{(s(t_i),a(t_i),r(t_i),\tau_i,s(t_{i+1}))\sim D}[Q_{\psi}(s(t_i),a(t_i))- (\frac{1-e^{-\rho \tau_i}}{\rho} r(t_i)+e^{-\rho \tau_i} V_\phi(s(t_{i+1})))  ]^2,
\end{equation}
as that in \cite{haarnoja2018soft}, where
$\tau_i = t_{i+1}-t_{i}$ can be calculated from the real data in  replay buffer $D$. \emph{Remark} that when we store $s(t_i)$ into the replay buffer, we also store its corresponding timestamp $t_i$.  Then this learned $V_\phi$ of current policy $\pi$ is used in the following policy learning.

In practice, we use the second approach and find its performance is better. It maybe because that the learned model has bias and we'd better use the real data.

\subsection{Policy learning:}\label{app_section:policy_learning}

Recall we have already obtained the NHPI model and transition model $g$. 

In particular, we replace the transition model $P(s'|a,s)$ by the learned transition model $g(s,a,\epsilon(s))$. Since we do the reparameterization, the randomness comes from the $\epsilon$ \cite{kingma2013auto}. Thus, the previous expectation in the Bellman equation $\mathbb{E}_{s'}(\cdot)$ (i.e., $\sum_{s'}P(s'|s,a)(\cdot)$) is now $\mathbb{E}_\epsilon (\cdot)$. We can sample over $\epsilon$ as that in \cite{kingma2013auto}. For the reward, we do similar things. We also replace the action by the policy, value function by learned one in above section.  Thus agent actually do the planning in the imaginary data set.

\begin{equation}
    \begin{split}
       \mathbb{E}_{\epsilon,\xi}[\int_{0}^{\infty} \int_{0}^{t}e^{-\rho\tau}\kappa\big(s(\tau), \pi_{\theta}(s(\tau)),\xi(s(\tau))\big)& d\tau dF(t|s,a,s')\\
     +  &\int_{0}^{\infty} e^{-\rho t} V_\phi\bigg(g\big(s(t), \pi_\theta(s(t)),\epsilon(s(t))  \big) \bigg)dF(t|s,a,s')]. 
    \end{split}
\end{equation}

There are several ways to do the sampling over the \emph{time interval}, i.e., the sampling according to $dF(t|s,a,s')$. Here we present what we did in the experiment.  In particular, we use the  dynamic model(NHPI) and the current policy to simulate the next time stamp $t_{i+1}$ with thinning algorithm. Then we calculate the time interval $\tau_i$ (Note it may not be same with  $\tau_i$ in the replay buffer, since the action sampled from the policy is not the same with occurred one ). Following the similar derivation in the value function learning,  one step rollout of the Bellman equation is    $$ V_{\phi,\theta}(s(t_i)):=(1-e^{-\rho\tau_i})  \kappa\big(s(t_i), \pi_{\theta}(s(t_i)),\xi(s(t_i))\big) /\rho
    + e^{-\rho \tau_i}V_\phi\bigg(g\big(s(t_i), \pi_\theta(s(t_i)),\epsilon(s(t_i))  \big) \bigg).$$
In practice, we randomly sample a  chunk of  $s(t_i)$ from the hidden states of NHPI with the input  trajectory from the replay buffer $D_{tr}$ or directly randomly sample $s(t_i)$ from $D$. For instance, in the former case, $i$ can be from $l$ to $l+H$, where $l$ is a random index. We update policy using gradient ascent of above equation. 

$$\theta \leftarrow \theta + \alpha \mathbb{E}_{s\sim D_{tr}}\frac{\partial V_{\phi,\theta}(s)}{\partial \theta} $$
    
% where we replace the transition, reward, and value function by the learned ones. Specifically, $r$ is replaced by $\kappa$, $s(t_{i+1})$ by $g\big(s(t_i), \pi_\theta(s(t_i)),\epsilon(s(t_i))  \big) $. 

We can also rollout the Bellman equation with several steps as Dreamer rather than one step. However, it may introduce more model bias and  we leave it as future work.

% The second one is simple. It uses the true time interval $\tau_i$ in replay buffer. Recall that given a true trajectory and timestamp $t_i$, we know the next timestamp $t_{i+1}$ from the true trajectory. This approach may omit the difference between the action pick by the policy and action from the replay buffer. In practice, it is slightly worse than the first one.

\section{Experiment Setting}\label{app_section:exp_setting}

In this section, we present the details of the experiment deferred from the main text. Event type number K, action num, and maximum simulation length is introduced in the Table \ref{tab:parameter}.  The event with type $k$ is encoded as a one-hot vector where $k$-th entry is one. Since $A\subseteq \mathcal{K}$, the action is encoded in the same way. The zero vector in actions means that we choose to do nothing.

\subsection{Synthetic experiment setting}

\begin{figure}
    \centering
    \subfloat[SI  setting]{\includegraphics[width= 0.5\textwidth]{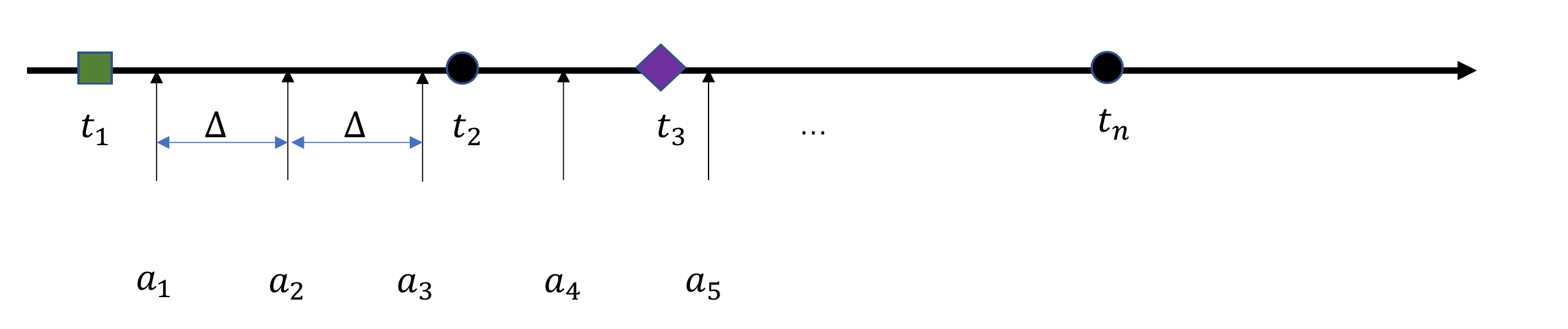}}
    \subfloat[USI setting]{\includegraphics[width= 0.5\textwidth]{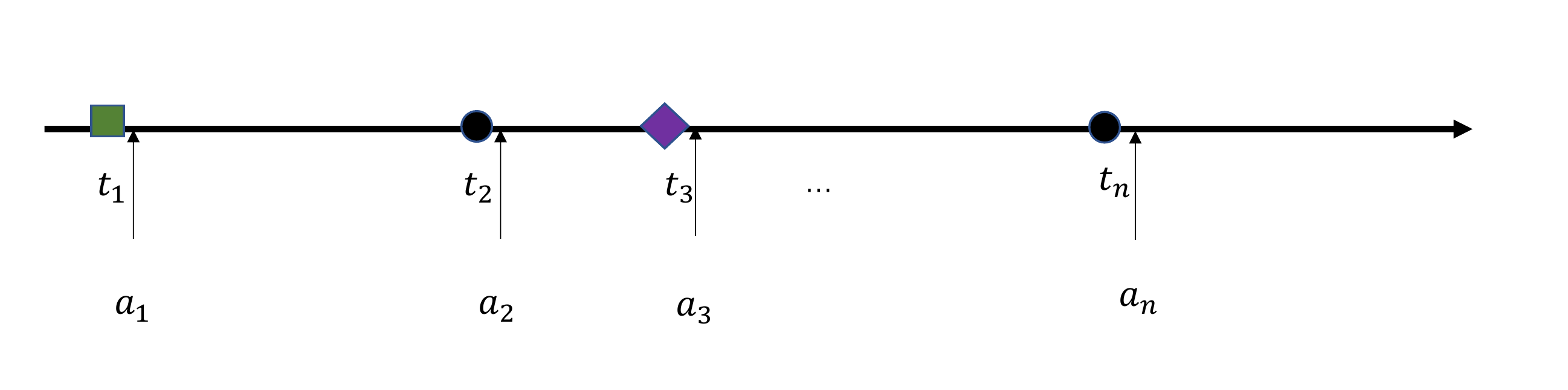}}
    \caption{The left panel shows the SI  setting. The right panel show  the USI  setting. }
    \label{fig:synthetic_setting}
\end{figure}

Figure \ref{fig:synthetic_setting} (a) demonstrates the SI  setting, where the agent can apply the action at evenly discretized intervals regardless when the event happens. In the USI  setting (b), the agent can only apply the action after events' occurrence.

In the experiment with K=8 event types, we only seek to control the intensity of the first  four event types, i.e., fake news.  $ \lambda_{target}$ for  these four event types are 0.3,0.3,0.3,0.3, respectively. The last four event types correspond to actions, i.e., valid news in our experiment. The reward is defined in the main text $r_t = -\| \bm{\lambda}_{target} - \bm{\lambda}_t \|^2 - 0.1*\text{cost}(a_t)$.  Although we do not care the intensity of these action events, they can lead to cost, since we have a cost of $1$ once we apply a action.

In the experiment with 16 events, we also have similar settings and we want to control the intensity of the first  eight event types. Their target intensity are $0.4,0.3,0.2,0.1,0.4,0.3,0.2,0.1,$ respectively.

\subsection{Real world application}\label{app:real-world}
% We implement real-world simulation on Stack-Overflow and Retweet dataset. 

% The Stack-Overflow is a popular website that the users could use post to ask questions or seek advises in various technical areas. Other users could answer the questions freely and the answer would be rated or awarded by other views. The whole dataset records the user awarding information along the two years time horizon with totally 6633 sequences. There are totally $U=22$ events in the datasets which indicate different types of positive award given by other users. In the simulation, we use the last half of events as the action candidate events to control the group awarding behaviour, which means that the agent should learn a control policy to induce specific awarding pattern by manipulating the action events. 

% The Retweet dataset records the user retweet behaviors using \textit{Twitter}, which is a social application used for information posting and sharing. This dataset contains totally 24000 retweet sequences with $U=3$ events. Each event indicates the post retweet behavior and differentiated by the size of followers. In the simulation, we use the last two events as the action candidate events. The agent would learn a strategy to control group retweet behaviours, which means the agent should mitigate or promote the candidate events to reach the predefined spread effect.

In the problems of  smart broadcasting and improving the engagement,  to build simulators( the feedback model of followers  or users), we modify the state-of-the-art neural Hawkes process model-THP \citep{zuo2020transformer} to include the exogenous intervention input following the work in \cite{ding2020traffic}. In particular, we follow the open source code from the authors to construct our simulator but with an additional input of the exogenous action \footnote{https://github.com/SimiaoZuo/Transformer-Hawkes-Process}.  For each simulation, we reset the environment by randomly sample one trajectory from the dataset,   and use first 10 events in the trajectory as the initial state for the generator. Then, we apply the action to the simulator. The simulator receives the influence of the action and then uses  the thinning algorithm \citep{lewis1979simulation,mei2017neural} to sample the next event and timestamps. The parameters of constructing the simulation has been introduced in the Table \ref{tab:parameter}. 

 In the smart broadcasting, we use the  Retweet dataset \cite{zhao2015seismic} to  fit a model of follower. The Retweet dataset records the user retweet behaviors using \textit{Twitter}, which is a social application used for information posting and sharing. This dataset contains totally 24000 retweet sequences with K=3 event types.  These event types correspond to observation event, and the rewards for them are -1,-0.2, -0.1, respectively. It says that  the occurrence of type 1 will cause a reward of -1, and occurrence of type 2 will incur a reward of -0.2, while occurrence of type 3 would result in a reward of -0.1.  This reward pattern means that the users may want to steer the activity with type 3 of her followers. We also use these three events as the action events, and the cost of each action event is -0.05.

In the problem of improving the engagement, we use the dataset of the Stack-Overflow to fit a feedback model of user.  The whole dataset records the user awarding information along the two years time horizon with totally 6633 sequences. There are totally K=22 event type. The first eleven type consists of our observation . The corresponding rewards of this event are  (-1.5,  -1.2,  -1.1,  -1.2,  -1.4,  -1.6,  -1.7,  -1.8 , -0.1, -0.1, -0.1 ) respectively. The other eleven event are our action event (e.g., good answer). The cost of each action event is -0.05.

\begin{table*}[h]
\centering
\begin{tabular}{ |c|c|c|c|c|  }
\hline
Environment & event number  & Action & Maximum time \\
Name & K & number & horizon  \\
\hline
8-event SI  setting & 8 & 4 & 100 \\
\hline
8-event USI   setting & 8 & 4 & 100  \\
\hline
16-event SI   setting & 16 & 8 & 100  \\
\hline
16-event USI setting & 16 & 8 & 100  \\
\hline
Smart broadcasting & 3 & 3 & 100 \\
\hline
Improving engagements & 22 & 11 & 100 \\
\hline
\end{tabular}
\caption{ Event number $K$, action number , maximum time horizon for each simulation environment implemented in the simulation experiment.}
\label{tab:parameter}
\end{table*}

\section{Settings of baselines and hyperparameters }\label{app_section:hyper}

The inputs of baselines are also the event sequence $\{(t_j, k_j ,a_j)\}_{j=1,..,L}$. The event with type $k$ is encoded as a one hot vector where the $k$-th entry is one.. Since $A\subseteq \mathcal{K}$, the action is encoded in the same way. The zero vector (in actions) means that we choose to do nothing. The timestamp is encoded as \eqref{app_equ:time_encoding}, which is then
fed into transformer as that in our NHPI model. Notice that these baselines (except Tpprl) treat the timestamp as a usual feature, and
therefore they are formulated as MDP problem. In tpprl, the agent is allowed to impose intervention at any time, where the intervention time is characterized by  the intensity function which is one element of the action. To fit our experimental setting, we fix the intervention time at $t_i$ in tpprl and   sample the event type from the softmax function.

In the following, we introduce the hyperparameter setting of our proposed model in different simulation and the maximum interaction with the environment is $150$K for all experiments.

The hyperparameter is determined by grid searching  for both our algorithms and baselines.  The number of hidden units is selected from the set $\{64,128,256\}$, the learning rate is picked from the set $\{1e-4,3e-4,1e-3\}$ and attention dimension is selected from $\{16,32,64\}$.  The discount rate $\rho$ is selected from $\{0.01,0.1\}$. Notice we can transfer this value into the usual discount factor $\gamma$ in RL by $\gamma= e^{-\rho}$. The hyperparameter used in our algorithm is listed in Table \ref{tab:hyper}. For the baselines with Transformer structure,they the same parameters with those of the Transformer in our algorithm, i.e., attention layer, head number and attention dimension.  For the baseline with LSTM structure,  one layer with hidden size 64 is used.

\begin{table}[h]
  \begin{center}
    \begin{tabular}{|c|c|c|c|c|c|c|}
      \hline
        & 8 SI  & 8 USI & 16 SI  & 16 USI & Stack-Overflow & Retweet\\
      \hline
      \tabincell{c}{ Maximum step \\ (interactions with environment)} & \multicolumn{6}{c|}{150k} \\
      \hline
      \tabincell{c}{Policy Learning Rate} & \multicolumn{6}{c|}{0.0001} \\
      \hline
      \tabincell{c}{ Value Learning Rate} & \multicolumn{2}{c|}{0.0001} & \multicolumn{2}{c|}{0.0003} & \multicolumn{2}{c|}{0.001}\\
      \hline
      \tabincell{c}{Model\\ Learning Rate} & \multicolumn{4}{c|}{0.0001} & \multicolumn{2}{c|}{0.001}\\
      \hline
      \tabincell{c}{Value and Policy\\ 2 layer MLP} & \multicolumn{2}{c|}{(128,64)} & \multicolumn{2}{c|}{(256,128)} & \multicolumn{2}{c|}{(256,256)} \\
      \hline
      \tabincell{c}{Transition \\2 layer MLP (output $\mu$, $\sigma$) } & \multicolumn{2}{c|}{(128,(64,64))} & \multicolumn{2}{c|}{(256,(128,128))}& \multicolumn{2}{c|}{(256, (128,128))}\\
      \hline
      \tabincell{c}{Reward \\2 layer MLP (output $\mu$, $\sigma$) } & \multicolumn{2}{c|}{(128,1)} & \multicolumn{2}{c|}{(256,1)}& \multicolumn{2}{c|}{(256, 1)}\\
      \hline
      \tabincell{c}{Attention \\ Head} & \multicolumn{2}{c|}{1} & \multicolumn{2}{c|}{2}& \multicolumn{2}{c|}{2}\\
      \hline
      \tabincell{c}{Attention\\ Layer} & \multicolumn{6}{c|}{2}\\
      \hline
      \tabincell{c}{Attention \\ Dimension $M_K, M_V$} & \multicolumn{2}{c|}{16} & \multicolumn{2}{c|}{32}& \multicolumn{2}{c|}{64}\\
      \hline
      \tabincell{c}{$\rho$ \\ (Discount Rate)} & \multicolumn{4}{c|}{0.01  $(e^{-0.01} \approx 0.99)$}   & \multicolumn{2}{c|}{0.1 $(e^{-0.1} \approx 0.905)$}\\
      \hline
     \tabincell{c}{Embedding dimension $M$} & \multicolumn{6}{c|}{64}\\
      \hline  
      \tabincell{c}{$w_m$} & \multicolumn{6}{c|}{ $\frac{1}{10000^{(m-1)/M}} $}\\
      \hline
    \end{tabular}
    \caption{The hyper-parameter used in our proposed method for each simulation environment. The number in policy, value, and model network architecture indicate the size of hidden units in each layer of MLP structure. The ReLu activation function is implemented in all architectures.}
    \label{tab:hyper}
  \end{center}
\end{table}

\section{Computational resource}\label{app_section:resourse}

We train our algorithm and all baselines with two Tesla V100 GPUs.

\section{Ablation Study}\label{app:ablation}
\begin{figure*}
\captionsetup[subfloat]{farskip=2pt,captionskip=1pt}
\centering

      \subfloat[SMDPs ablation (a)]{\includegraphics[width=.32\textwidth]{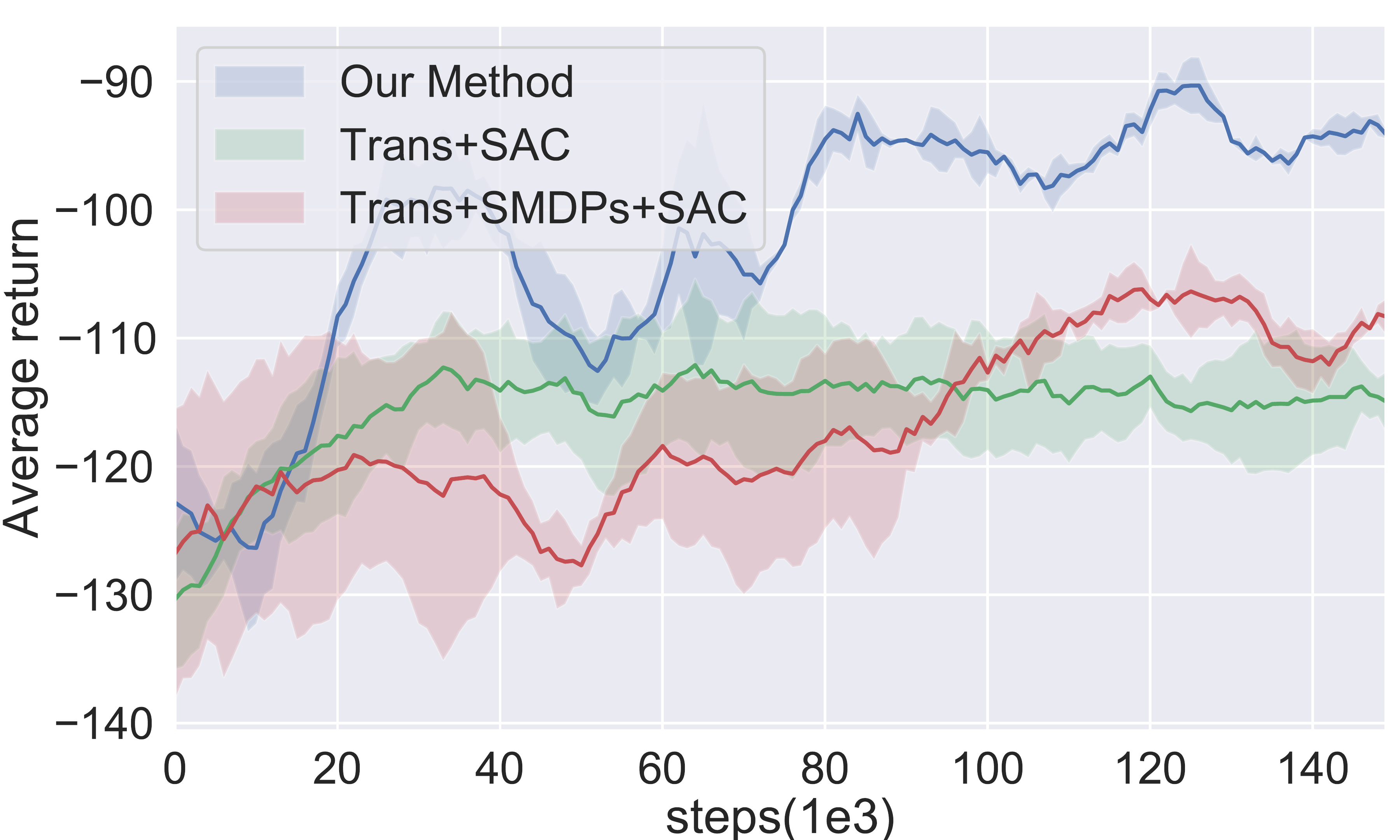}}\hspace{0.01em}
    \subfloat[SMDPs ablation (b)]{\includegraphics[width=.32\textwidth]{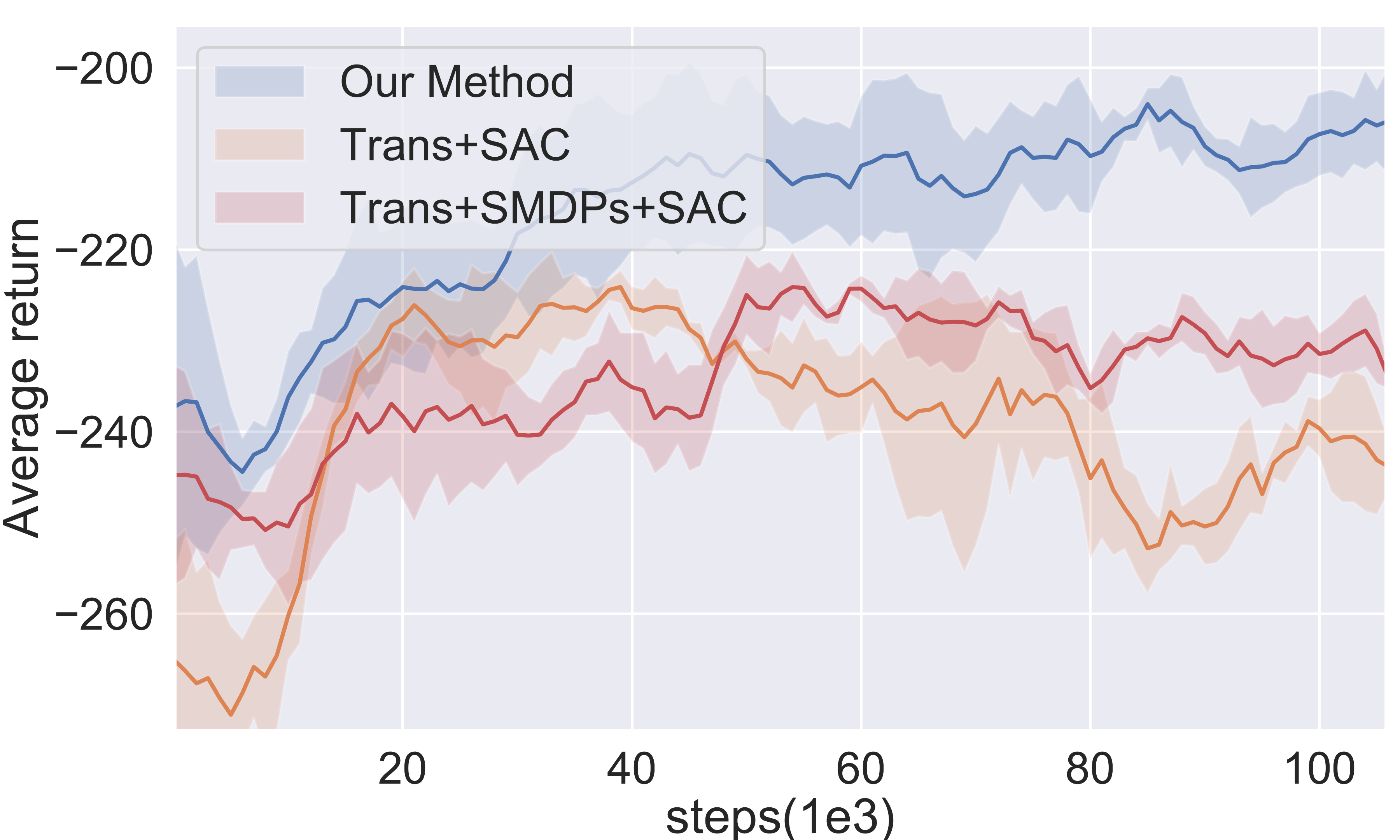}}\hspace{0.01em}
    \subfloat[MLP ablation]{\includegraphics[width=.32\textwidth]{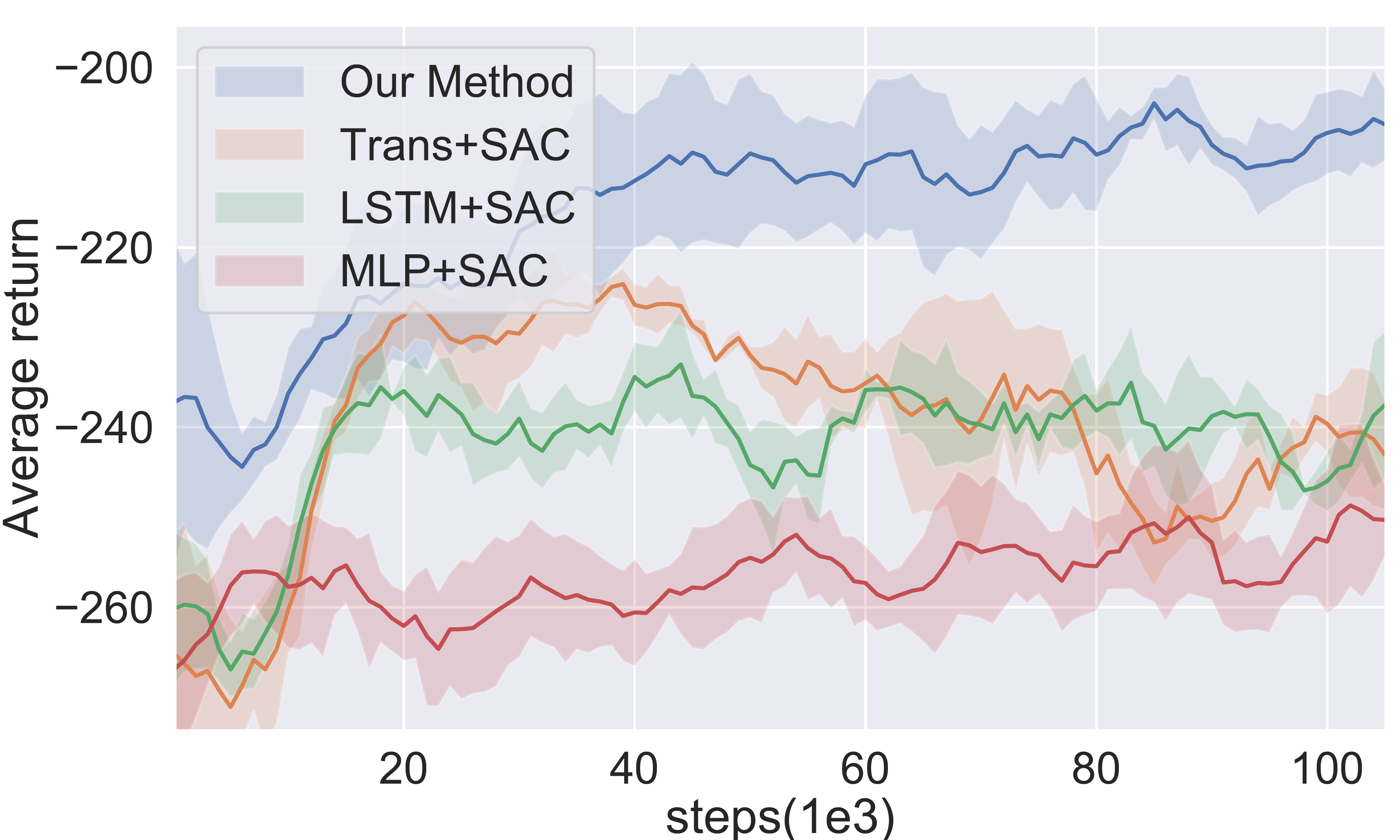}}\hspace{0.01em}
   \caption{ The test results of our ablation study. The x-axis is the training step and y-axis is the cumulative reward. Ten trials are tested in each simulation using various random seeds and initialized parameters. The average return is reported as the solid line and the standard deviation is reported as shaded region. 
}
   \label{fig:ablation}
\end{figure*}

In this ablation study, we would like to answer two questions: 1. what is the performance of the baseline if we adapt them into SMDPs version. 2. What is the performance of traditional RL algorithm with MLP structure under event-driven setting?

First, we adapt SAC algorithm  into it SMDPs counterpart and test it in the 8 and 16 USI setting, for which we replace the target function of $Q$ in SAC by  $\frac{1-e^{-\rho \tau_i}}{\rho} r(t_i)+e^{-\rho \tau_i} Q(s_{i+1},a_{i+1})$, where $\tau_i$ is the time interval. The test result is shown in  Figure \ref{fig:ablation}(a,b) where we can see its SMDP version can improve the performance over  vanilla version. Especially when the length of time interval is unequal in the USI setting simulation. However, we can still observe our proposed method having much better performance, which is contributed by other perspectives such as hidden dynamics from our algorithm.  

Then we test the performance of SAC algorithm with usual multilayer perceptron structure (2-layer MLP with hidden size of (256,128)) under event-driven setting.   The test is performed in 16-event USI simulation and the test result is shown in the Figure \ref{fig:ablation} c. We can observe that the model is barely working under this setting, which is well-expected, since the environment is partially observable.

\section{Limitation and Future Work}
One possible limitation of our work is that we use the dynamic model to generate imaginary data to train the agent, which may cause the model bias in RL. In the future work, we would combine model-based and model-free method together to mitigate such issue

\end{document}